\RequirePackage{fix-cm}
\documentclass{svjour3}                     
\smartqed  
\usepackage[acronym,toc,nopostdot,style=super,nonumberlist]{glossaries}
\usepackage[numbers]{natbib}
\usepackage{graphicx}
\usepackage{multirow}
\usepackage{algpseudocode}
\usepackage{algorithm}
\usepackage{booktabs} 
\usepackage{amsmath} 
\usepackage{xcolor}
\usepackage{subcaption}
\usepackage{comment} 
\usepackage[section]{placeins} 

\graphicspath{{Images/}}
\journalname{}

\def\makeheadbox{\relax}
\begin{document}

\newacronym{SER}{SER}{Scalarised Expected Return}
\newacronym{ESR}{ESR}{Expected Scalarised Return}
\newacronym{RL}{RL}{reinforcement learning}
\newacronym{MORL}{MORL}{multi-objective reinforcement learning}
\newacronym{MDP}{MDP}{Markov decision process}
\newacronym{MOMDP}{MOMDP}{multi-objective Markov decision process}
\newacronym{MDPs}{MDPs}{Markov decision processes}
\newacronym{MOMDPs}{MOMDPs}{multi-objective Markov decision processes}
\newacronym{HRL}{HRL}{Hierarchical reinforcement learning}
\newacronym{IRL}{IRL}{Inverse reinforcement learning}
\newacronym{SMDPs}{SMDPs}{semi-Markov decision processes}
\newacronym{TLO}{TLO}{thresholded lexicographic ordering}
\newacronym{DRL}{DRL}{distributional reinforcement learning}


\title{An Empirical Investigation of Value-Based Multi-objective Reinforcement Learning for Stochastic Environments}


\titlerunning{An Empirical Investigation of Value-Based MORL for Stochastic Environments}        

\author{Kewen Ding \and
        Peter Vamplew \and
        Cameron Foale \and
        Richard Dazeley
}


\institute{
            K. Ding \at
              Federation University Australia \\
              \email{k.ding@federation.edu.au} 
              \and
           P. Vamplew \at
              Federation University Australia \\
    \email{p.vamplew@federation.edu.au} 
              \and
           C. Foale \at
              Federation University Australia\\
              \email{c.foale@federation.edu.au}
              \and
           R.Dazeley \at 
              Deakin University\\
              \email{richard.dazeley@deakin.edu.au}
}


\maketitle

\begin{abstract}
One common approach to solve multi-objective reinforcement learning (MORL) problems is to extend conventional Q-learning by using vector Q-values in combination with a utility function. 
However issues  can arise with this approach in the context of stochastic environments, particularly when optimising for the Scalarised Expected Reward (SER) criterion. This paper extends prior research, providing a detailed examination of the factors influencing the frequency with which value-based MORL Q-learning algorithms learn the SER-optimal policy for an environment with stochastic state transitions. We empirically examine several variations of the core multi-objective Q-learning algorithm as well as reward engineering approaches, and demonstrate the limitations of these methods. In particular, we highlight the critical impact of the \emph{noisy Q-value estimates} issue on the stability and convergence of these algorithms. 
\end{abstract}

\section{Introduction}

The goal of \acrfull{MORL} is to expand the generality of \acrfull{RL} methods to enable them to work for problems with multiple conflicting objectives \cite{Hayes2022practicalguide, roijers2013survey}. Traditional \acrshort{RL} normally assumes that the environment is a \acrfull{MDP} in which the agent will receive a scalar reward after performing each action, and the goal is to learn a policy that maximises a single long-term return based on those rewards \cite{sutton2018reinforcement}. In contrast, \acrshort{MORL} works with \acrfull{MOMDPs}, where the reward values are vectors, with each element in the vector corresponding to a different objective. Using vector rewards overcomes the limitations of scalar rewards \cite{vamplew2022scalar} but also creates a number of new algorithmic challenges.

One of the most common approaches used so far in the MORL literature is to extend standard scalar RL algorithms such as Q-learning or Deep Q-Networks to handle vector rewards. We will cover the details of how this is achieved in Section \ref{sec:moq}. While this method has been successfully applied, it has also been demonstrated to have some significant shortcomings, particularly in the context of environments with stochastic rewards and/or state transitions \cite{VamplewEnvironmental2022}.

This paper provides a more detailed exploration of the issues identified by \cite{VamplewEnvironmental2022}. We explore various methods by which the issues caused by the local decision-making aspect of Q-learning might be solved or ameliorated, including changes in reward design, as well as algorithm modifications. In addition we examine the extent to which the \emph {noisy Q-value estimates} issue is the main factor impeding the ability of value-based MORL methods to converge to optimal solutions in stochastic environments.

Section 2 provides the required background, giving a general introduction to MORL and MOMDPs, as well as more specific detail on the issues faced by value-based MORL algorithms in environments with stochastic state dynamics. Section 3 provides an overview of the experimental methodology, including the Space Traders MOMDP which we will be used as a benchmark. The following four sections report and discuss experimental results from four different approaches (baseline multiobjective Q-learning, reward engineering, an extension of MO Q-learning to incorporate global statistics, and MO Q-learning using policy options). The paper concludes in Section 8 with an overview of the findings, and suggestions for future work to address the task of learning optimal policies for stochastic MOMDPs. 
\color{black}

\section{Background}

\subsection{Multi-objective reinforcement learning}
The basic multi-objective sequential decision problem can be formalised as a \acrfull{MOMDP}. It is represented by the tuple $\langle S, A, T, \mu, \gamma, \bf{R} \rangle$ where:
\begin{itemize}
    \item $S$ is a finite set of states
    \item $A$ is a finite set of actions
    \item $T: S\times A\times S\rightarrow[0,1]$ is a state transition function
    \item $\mu: S\rightarrow[0,1]$ is a probability distribution over initial states
    \item $\gamma\in[0,1)$ is a discount factor 
    \item $\bf{R}$ $: S\times A\times S\rightarrow R^d$ is a vector-valued reward function which defines the immediate reward for each of the $d\geq2$ objectives.
\end{itemize}

So the main difference between a single-objective MDP and a MOMDP is the vector-valued reward function $\bf{R}$, which specifies a numeric reward for each of the considered objectives. The length of the reward vector is equal to the number of objectives. The generalisation of RL to include vector rewards introduces a number of additional issues. Here we will focus on those which were of direct relevance to this study; for a broader overview of MORL we recommend \cite{Hayes2022practicalguide,roijers2013survey}.

\subsubsection{Action selection and scalarisation}
The most obvious issue is that the optimal policy is less clear because there may be multiple optimal policies (in terms of Pareto optimality). Therefore, \acrshort{MORL} requires some approach for ordering those vector values.

There has been a trend in recent literature to adopt a utility-based approach as proposed by \cite{roijers2013survey}. This approach utilises domain knowledge to define a utility function which captures the preferences of the user. In value-based MORL, the utility (aka scalarisation) function is used to determine an ordering of the Q-values for the actions available at each state, in order to determine the action which is optimal with respect to the user's utility. These functions can be broadly divided into two categories which are linear and monotonically increasing nonlinear functions. 

Linear scalarisation is straightforward to implement, as it converts the MOMDP to an equivalent MDP \cite{roijers2013survey}. However, it also suffers from a fundamental disadvantage that it is incapable of finding solutions which lie in concave regions of the Pareto front \cite{2008On}. Also in some situations, a linear scalarisation function is not sufficient to handle all types of user preferences (for example, MORL approaches to fairness in multi-user systems use nonlinear functions such as the Nash Social Welfare function or the Generalised Gini Index \cite{siddique2020learning,fan2022welfare}).

Therefore, monotonically increasing (nonlinear) scalarisation functions have been introduced (e.g. \cite{2013Scalarized}). These adhere to the constraint that if a policy increases for one or more of the objectives without decreasing any of the other objectives, then the scalarized value also increases \cite{Hayes2022practicalguide}. One notable example is the thresholded lexicographic ordering (TLO) method which allows agent to select actions prioritised in one objective and meet specified thresholds on the remaining objectives \cite{gabor1998multi,issabekovEmpiricalComparisonTwo2012}.

Under nonlinear functions (such as TLO) the rewards are no longer additive which violates the usage of the Bellman equation for value-based methods \cite{roijers2013survey}. To address this, for value-based MORL both action selection and Q-values must be conditioned on the current state as well as a summary of the history of prior rewards (see Algorithm \ref{algo:moql-expected} -- line 11 creates an \emph{augmented state} via a concatenation of the environmental state and the history of prior words, and Q-values and action selection are based off this augmented state).

\subsubsection{Single-policy vs multi-policy methods}

In single-objective RL the aim is to find a single, optimal policy. In contrast for MORL, an algorithm may need to find a single or multiple policies depending on whether or not the user is able to provide the utility function prior to the learning or planning phase. For example, if user already knows in advance their desired trade-off between each objective, then the utility function is known in advance and fixed. Therefore there is no need to learn multiple policies as the agent can simply find the optimal policy which maximises that utility. On the other hand, if the utility function can not be designed before the training or the preference could change over time, then the agent has to return a \emph{coverage set} of all potentially optimal policies. The user will then select from this set to determine which policy will be used in a particular episode. 

If we consider the scenario of planning a trip as an example, the traveller  may or may not know the exact preferences about getting to the destination in terms of when to get there and how much the traveller is willing to spend on this journey. So in this case, the algorithm needs to learn all non-dominated policies. However, if the traveller has a preference about how long they can take to arrive at their destination or there is a certain budget associated with this trip, then a single policy will be enough to satisfy their preferences.

\subsubsection{Scalarised Expected Returns versus Expected Scalarised Returns}

According to \citeauthor{roijers2013survey}\cite{roijers2013survey}, there are two distinct optimisation criteria compared with just a single possible criteria in conventional RL\footnote{The conventional single-objective RL does not use a scalarisation function, and so the ESR and SER criteria are the same in this context.}. The first one is \acrfull{ESR}. In this approach, the agent aims to maximise the expected value which is first scalarised by the utility function, as shown below (Eq \ref{eq:ESR}) where $w$ is the parameter
vector for utility function $f$, $r_k$ is the vector reward on time-step $k$, and $\gamma$ is the discounting factor
\begin{equation}
\label{eq:ESR}
V^\pi_{\bf w}(s) =  E[f( \sum^\infty_{k=0} \gamma^k \mathbf{r}_k,{\bf w})\ | \ \pi, s_0 = s)
\end{equation}
ESR is the appropriate criteria for problems where the aim is to maximise the expected outcome within each individual episode. A good example is searching for a treatment plan for a patient, where there is a trade-off between cure and negative side-effect. Each patient would only care about their own individual outcome instead of the overall average. 

The second criteria is \acrfull{SER} which estimates the expected rewards per episode and then maximises the scalarised expected return, as shown in (Eq \ref{eq:SER})
\begin{equation}
\label{eq:SER}
V^\pi_{\bf w}(s) = f( {\bf V}^\pi(s), {\bf w}) =  f(E[\sum^\infty_{k=0} \gamma^k \mathbf{r}_k \ | \ \pi, s_0 = s], {\bf w})
\end{equation}
So \acrshort{SER} formulation is used for achieving the optimal utility considered over multiple executions. Continuing with the travel example, the employee wants to cut down on the amount of time spent travelling to work each day. Travelling by car would be the good option on average, although there may be rare days on which it is considerably slower due to an accident. 

\subsection{Multiobjective Q-learning and stochastic environments}\label{sec:moq}

In contrast to much of the prior work on MORL which has used deterministic environments, \cite{VamplewEnvironmental2022} examined the behaviour of multi-objective Q-learning in stochastic environments. They demonstrated that in order to find the SER-optimal policy for problems with stochastic rewards and a non-linear utility function, the MOQ-learning algorithm needs action selection and Q-values to be conditioned on the summed expected rewards in the current episode, rather than the summed actual rewards (see Lines 15-18 in Algorithm \ref{algo:moql-expected}).

\begin{algorithm}[hbt!]
  \caption{Multi-objective Q($\lambda$) using accumulated expected reward as an approach to finding deterministic policies for the SER context (\cite{VamplewEnvironmental2022}).}
  \label{algo:moql-expected}

  \begin{algorithmic}[1]
    \Statex input: learning rate $\alpha$, discounting term $\gamma$, eligibility trace decay term $\lambda$, number of objectives $n$, action-selection utility function $f$ and any associated parameters
    \For {all states $s$, actions $a$ and objectives $o$}
    	\State initialise $Q_o(s,a)$
    	\State initialise $I_o(s,a)$ \Comment{estimated immediate (single-step) reward}
    \EndFor
    \For {each episode} 
        \For {all states $s$ and actions $a$}
    		\State $e(s,a)$=0
        \EndFor
        \State sums of prior expected rewards $P_o$ = 0, for all $o$ in 1..$n$
    	\State observe initial state $s_t$
    	\State $s_t$ = $(s_t,P)$ \Comment{create augmented state}
        \State \parbox[t]{\dimexpr\linewidth-\algorithmicindent\relax}{%
            \setlength{\hangindent}{\algorithmicindent}%
            select $a_t$ from an exploratory policy derived using $f(Q(s))$
        }\strut
        \For {each step of the episode}
 			\State execute $a_t$, observe $s_{t+1}$ and reward $R_t$
 			\State update $I(s_t,a_t)$ based on $R_t$
 			\State $P = P + I(s_t,a_t)$
 			\State $s_{t+1}$ = $(s_{t+1},P)$ \Comment{create augmented state}
 			\State $U(s_{t+1}) = Q(s_{t+1}) + P$ \Comment{create value vector}
 			\State select $a^*$ from a greedy policy derived using $f(U(s_{t+1}))$
  			\State select $a^\prime$ from an exploratory policy derived using $f(U(s_{t+1}))$
            \State $\delta = R_t + \gamma Q(s_{t+1},a^*) - Q(s_t,a_t)$
            \State $e(s_t,a_t)$ = 1
            \For {each state $s$ and action $a$}
            	\State $Q(s,a) = Q(s,a) + \alpha\delta e(s,a)$
                 \If {$a^\prime = a^*$}
                 	\State $e(s,a) = \gamma \lambda e(s,a)$
                 \EndIf
            \EndFor
            \State $s_t = s_{t+1}, a_t = a^\prime$
    	\EndFor
    \EndFor
  \end{algorithmic}
\end{algorithm}

\cite{VamplewEnvironmental2022} also identified that existing value-based model-free MORL methods may fail to find the \acrshort{SER} optimal policy in environments with stochastic state transitions. Under this type of environment, following the same policy may result in different trajectories and rewards in each episode time. Since the \acrshort{SER} criteria aims to achieve the optimal utility over multiple executions,  the overall policy in order to meet that constraints depends on the probability with which each trajectory is encountered. Therefore determining the correct action to select at each possible trajectory requires the agent to also consider the returns received in every other trajectory in combination with the probability of that trajectory having been followed \cite{bryceProbabilisticPlanningMultiobjective2007}. This requirement is incompatible with standard value-based model-free methods like Q-learning, where it is assumed that the best action can be fully determined from the local information available to the agent at the current state. Augmenting that state information with the sum of expected rewards as in Algorithm \ref{algo:moql-expected} is insufficient as this still only provides information about the trajectory which has been followed in this episode, rather than all possible trajectories that agent might be able to reach under this same policy.

A second issue, identified by both \cite{vamplew2021potential} and \cite{VamplewEnvironmental2022}, is the problem of \emph{noisy Q-value estimates}. The current policy of the agent is determined by the Q-values, which estimate the value of each action in the current state. Small errors in these estimates may lead to the selection of an alternative action. When this arises in the context of single-objective RL, the potential impact is low, as the incorrectly selected action must have a similar value as the greedy action (within the bounds of the noise in the estimates). However in the context of MORL, this impact can be much larger as two actions with very different reward vectors may have similar scalarised values. This is particularly true for highly non-linear functions like TLO. Here a small change in the estimated value of the thresholded objective for an action can lead to it incorrectly being regarded as now satisfying the threshold (or vice-versa). Hence small amounts of noise may have a large impact on the actual reward vector received. While this problem is not specific to stochastic environments, it will be more evident in this context as the variation in future returns due to the stochasticity will in itself result in greater variation in the Q-value estimates.

One of the main aims of this paper is to examine the extent to which these factors interfere with learning SER-optimal policies for stochastic environments, and to explore possible approaches for addressing these issues, while remaining within the overall model-free value-based MORL framework.

\section{Experimental methodology}

This paper investigates several alternative approaches to address the issues arising in value-based SER-optimal MORL. We provide here an overview of the aspects of the experimental methodology which were common across all experiments, while later sections will provide details of the individual approaches and any specific experimental modifications required during their evaluation.

\subsection{Approaches tested}
The experiments evaluate four \acrshort{MORL} methods, including a baseline method. 
\begin{itemize}
\item Baseline approach - This is the basic MOQ-learning with expected accumulated reward shown in Algorithm \ref{algo:moql-expected}. This was used to replicate the original results of \cite{VamplewEnvironmental2022} to serve as a baseline for evaluating the performance of the other approaches.
\item Reward engineering approach - This approach modifies the design of the reward signal, while continuing to use the baseline algorithm.
\item Global statistic approach - Here we introduce a novel heuristic algorithm which includes global statistical information during action selection in an attempt to address the issues caused by purely local decision-making.
\item Options-based approach - This approach uses the concept of an option as a `meta-action' which determines the action selection over multiple time-steps compared with single time-step in the baseline method.
\end{itemize}

In parallel with these experiments, we also investigate the impact of the noisy estimates issue on the performance of these algorithms. This is achieved by decaying the learning rate of MOQ-learning from its initial value to zero over the training period. This ensures that the Q-values stabilise around their true values, and allows us to examine the impact this has on each approach's ability to correctly identify the SER-optimal policy.

\subsection{Performance Measure}
The focus of this paper is on evaluating how effectively value-based MORL can identify the SER-optimal policy for a MOMDP with stochastic state transitions. Therefore our key metric is the frequency with which each approach converges to the desired optimal policy. Each approach was executed for twenty trials, and we measure how many of these trials result in a final greedy policy which is SER-optimal.

For each of the approaches implemented in this research, the following data was collected for each of the twenty trials in each experiment.
\begin{itemize}
\item The reward that is collected by the agent during 20,000 episodes of training
\item For each episode, the greedy policy according to the agent's current Q-values (note: this policy was not necessarily followed during this episode, due to the inclusion of exploratory actions)
\item After training, the final greedy policy learnt by the agent. This was compared against the SER-optimal policy for the environment to determine whether the trial was a success or not.
\end{itemize}

\subsection{Space Trader Environment}
\begin{figure}[hbt!]
    \centering
    \includegraphics[width=12cm]{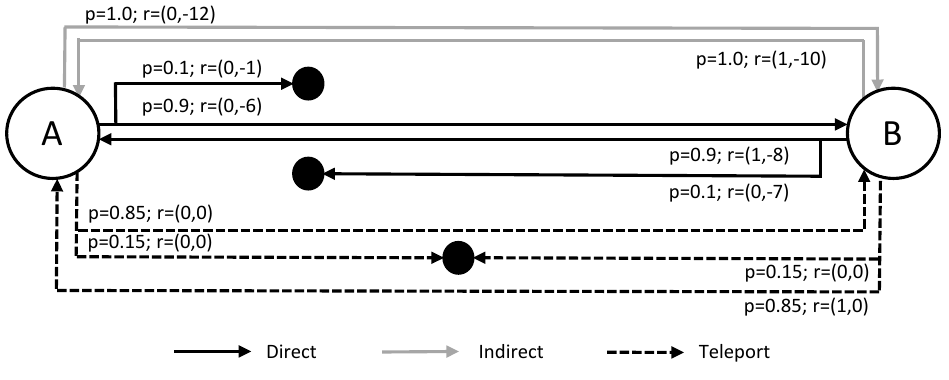}
    \caption[The original Space Traders Environment]
    {The Space Traders MOMDP. Solid black lines show the Direct actions, solid grey line show the Indirect actions, and dashed lines indicate Teleport actions. Solid black circles indicate terminal (failure) states \protect\cite{VamplewEnvironmental2022}}
    \label{fig:ST-O}
\end{figure}

The Space Traders shown in Fig \ref{fig:ST-O} was the environment used by \cite{VamplewEnvironmental2022} to identify the issues discussed in Section 2.2, and so it will form the basis for our experiments. It is a simple finite-horizon task with only two steps and it consists of two non-terminal states with three actions (direct, indirect and teleport) available to choose from each state. The agent starts from planet A (State A) and travels to planet B (state B) to deliver shipment and then returs back to planet A with the payment. The reward for each action consists of two parts. The first element is whether the agent successfully returned back to planet A. So the agent only receives 1 as reward on last successful action and 0 for all other actions,  including those which result in a terminating state corresponding to mission failure. The second element is a negative penalty which indicates how long this action takes to execute.

Table \ref{tab:Space-Traders-O1} shows the transition probabilities, immediate reward for each state action pairs and mean rewards as well. The reason for selecting Space Traders as the testing environment is because it is a relatively small environment. So, it is easy to list all of the nine possible deterministic policies which are shown in Table \ref{tab:Space-Traders-O2}. For these experiments we assume the goal of the agent is to minimise the time taken to complete the travel as well as having at least equal or above 88\% probability of successful completion (i.e. we are using TLO, with a threshold applied to the success objective). Under this utility function, the optimal policy is DI as it is the fastest policy which achieves a return of 0.88 or higher for the first objective.

\begin{table}[]
    \centering
\begin{tabular}{|l|l|l|l|l|l|}
\hline
State              & Action   & P(success) & \begin{tabular}[c]{@{}l@{}}Reward \\ on\\ success\end{tabular} & \begin{tabular}[c]{@{}l@{}}Reward \\ on\\ failure\end{tabular} & \begin{tabular}[c]{@{}l@{}}Mean\\ reward\end{tabular} \\ \hline
\multirow{3}{*}{A} & Indirect & 1.0        & (0,-12)                                                        & n/a                                                            & (0,-12)                                               \\[5pt] \cline{2-6} 
                   & Direct   & 0.9        & (0, -6)                                                        & (0, -1)                                                        & (0, -5.5)                                             \\[5pt] \cline{2-6} 
                   & Teleport & 0.85       & (0,0)                                                          & (0,0)                                                          & (0, 0)                                                \\[5pt] \hline
\multirow{3}{*}{B} & Indirect & 1.0        & (1, -10)                                                       & n/a                                                            & (1, -10)                                              \\[5pt] \cline{2-6} 
                   & Direct   & 0.9        & (1, -8)                                                        & (0, -7)                                                        & (0.9, -7.9)                                           \\[5pt] \cline{2-6} 
                   & Teleport & 0.85       & (1, 0)                                                         & (0, 0)                                                         & (0.85, 0)                                             \\[5pt] \hline
\end{tabular}
    \caption[The probability of success and reward values for each state-action pair in the Space Traders MOMDP]
    {The probability of success and reward values for each state-action pair in the Space Traders MOMDP \protect\cite{VamplewEnvironmental2022}}
    \label{tab:Space-Traders-O1}
\end{table}

\begin{table}[]
\centering
\begin{tabular}{|l|l|l|l|}
\hline
\begin{tabular}[c]{@{}l@{}}Policy\\ identifier\end{tabular} & \begin{tabular}[c]{@{}l@{}}Action in\\ state A\end{tabular} & \begin{tabular}[c]{@{}l@{}}Action in\\ state B\end{tabular} & Mean return     \\ \hline
II                                                          & Indirect                                                    & Indirect                                                    & (1, -22)        \\[5pt] \hline
ID                                                          & Indirect                                                    & Direct                                                      & (0.9, -19.9)    \\[5pt] \hline
IT                                                          & Indirect                                                    & Teleport                                                    & (0.85, -12)     \\[5pt] \hline
DI                                                          & Direct                                                      & Indirect                                                    & (0.9, -14.5)    \\[5pt] \hline
DD                                                          & Direct                                                      & Direct                                                      & (0.81, -12.61)  \\[5pt] \hline
DT                                                          & Direct                                                      & Teleport                                                    & (0.765, -5.5)   \\[5pt] \hline
TI                                                          & Teleport                                                    & Indirect                                                    & (0.85, -8.5)    \\[5pt] \hline
TD                                                          & Teleport                                                    & Direct                                                      & (0.765, -6.715) \\[5pt] \hline
TT                                                          & Teleport                                                    & Teleport                                                    & (0.7225, 0)     \\[5pt] \hline
\end{tabular}
    \caption[Mean return for nine deterministic policies in original Space Traders Environment]{The mean return for the nine available deterministic policies for the Space Traders Environment \protect\cite{VamplewEnvironmental2022}}
    \label{tab:Space-Traders-O2}
\end{table}

The hyperparameters used for experiments on the Space Traders environment can be found in Table \ref{tab:Space-Traders-O3}. These were kept the same across all experiments so as to facilitate fair comparison between the different approaches. For exploration we use a multi-objective variant of softmax (softmax-t) \cite{vamplew2017softmax}.

\begin{table}[]
\begin{tabular}{|l|l|l|l|l|l|l|}
\hline
Parameter & $\alpha$ & $\lambda$ & $\gamma$ & \begin{tabular}[c]{@{}l@{}}softmax-t \\initial \\temperature\end{tabular} & \begin{tabular}[c]{@{}l@{}}softmax-t \\ final\\temperature\end{tabular} & \begin{tabular}[c]{@{}l@{}}Number of episodes \\ per training run\end{tabular} \\ \hline
Value     & 0.01     & 0.95      & 1        & 10                                                                       & 2                                                                      & 20,000                                                                         \\ \hline
\end{tabular}
    \caption[Parameter table for Space Traders]{Hyperparameters used for experiments with the Space Traders environment.}
    \label{tab:Space-Traders-O3}
\end{table}

In some of the following experiments we will introduce variations of the original Space Traders environment in order to demonstrate that methods solving the original problem may fail under small changes in environmental or reward structure, illustrating that they do not provide a general solution to the problem of learning SER-optimal policies.

\subsection{Scope}
We have made several decisions to restrict the scope of this study, so as to focus on the specific issues of interest (solving SER-optimality for stochastic state transitions, and analysing the impact of noisy estimates on MOQ-learning). 

These issues can arise both in the context of single-policy and multi-policy MORL, but here we consider only single-policy approaches to simplify the analysis. Similarly we examine only a single choice of utility function -- thresholded lexicographic ordering was selected as it has been widely used in the MORL literature \cite{dornheim2022gtlo,gabor1998multi,hayes2020dynamic,issabekovEmpiricalComparisonTwo2012,jin2017multi,lian2023inkjet,tercan2022solving}, and has previously been shown to be particularly sensitive to noisy Q-value estimates \cite{VamplewEnvironmental2022}. Finally the simplicity of the Space Traders task allows us to use a tabular form of MOQ-learning, meaning that the noisiness of the estimates arises directly from the stochasticity of the environment. The problems identified in this study would be expected to be even more prevalent for Deep MORL methods, where the use of function approximation introduces an additional source of error for the Q-values.

\section{Experimental Results - Baseline MOQ-learning}\label{sec:baseline}
This experiment used the baseline MOQ-learning algorithm. The aim was to confirm the original findings of \cite{VamplewEnvironmental2022}, and provide a baseline for the later experiments.
Table \ref{tab:Space-Traders-Baseline} summarises the distribution of the final greedy policy learned over twenty independent training results.

\begin{table}[hbt!]
    \centering
    \begin{tabular}{@{}|c|c|c|c|c|@{}}
    \hline
    Policy & DI & ID & II & IT  \\ 
    \hline
    Baseline & 1 & 13 & 4  & 2    \\ \hline
    \end{tabular}%
    \caption[Twenty independent runs of the Algorithm \ref{algo:moql-expected}]{The final greedy policies learned in twenty independent runs of the baseline multi-objective Q-learning algorithm (Algorithm \ref{algo:moql-expected}) on the Space Traders environment}
    \label{tab:Space-Traders-Baseline}
\end{table}

The empirical results show that the desired optimal policy (DI) was not converged to in practice, with it being identified as the best policy in only one of twenty runs. This is comparable with the results reported for this method by \cite{VamplewEnvironmental2022}. They explained this behaviour by noting that regardless of which action the agent selected at state A, if state B is successfully reached, then a zero reward will have been received by the agent for the first objective. In other words, the accumulated expected reward for the first objective at state B is zero. Therefore, the choice of action at state B is purely based on the action values at that state. Now looking at the mean action values for state B which is reported in Table \ref{tab:Space-Traders-O1}, it can be seen that the teleport action will be eliminated because it fails to meet the threshold for the first objective, and the direct action will be preferred over indirect action as both meet the threshold, and direct action takes less time penalty in second objective. Therefore, the agent will choose the direct action at state B regardless of which action the agent selected at state A. As the result, this agent at state A will only consider Policy ID, DD and TD and only policy ID is above the threshold for the first objective if we look back the mean reward in table \ref{tab:Space-Traders-O2}. Therefore the agent converges to the sub-optimal policy ID in most trials. 

\begin{figure}[hbt!]
     \centering
     \begin{subfigure}[h]{0.45\textwidth}
         \centering
         \includegraphics[width=\textwidth]{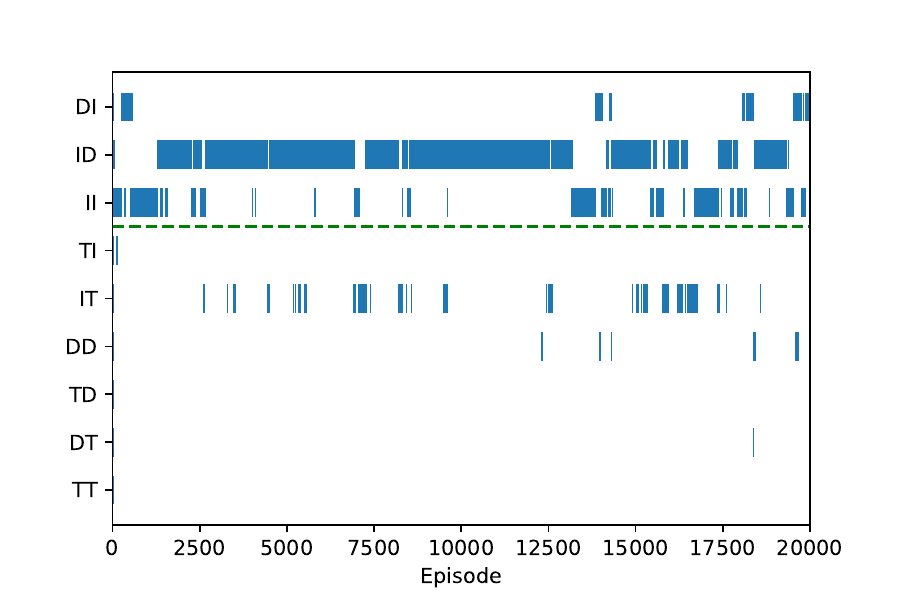}
         \caption{Final policy = DI}
         \label{fig:Baseline DI}
     \end{subfigure}
     \begin{subfigure}[h]{0.45\textwidth}
         \centering
         \includegraphics[width=\textwidth]{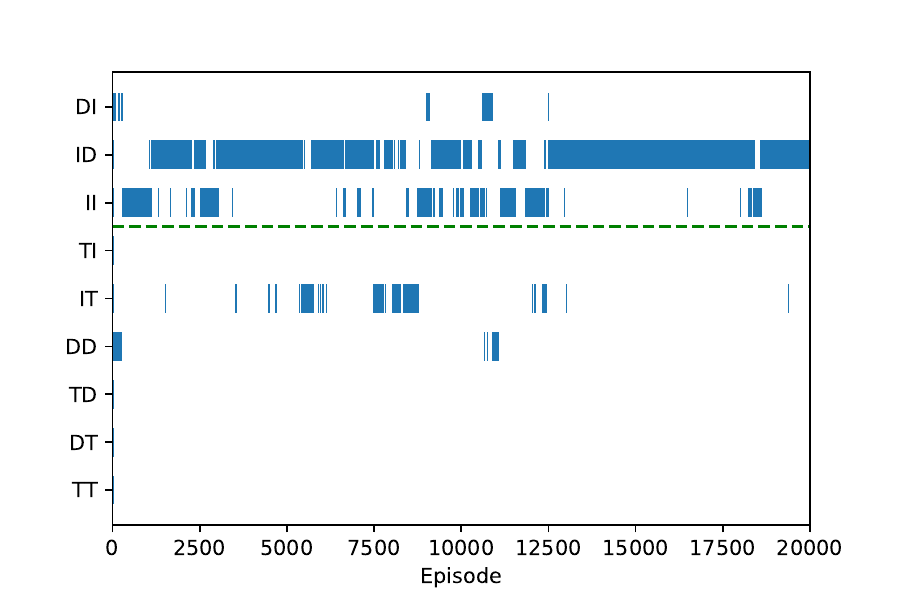}
         \caption{Final policy = ID}
         \label{fig:Baseline ID}
     \end{subfigure}
     \begin{subfigure}[h]{0.45\textwidth}
         \centering
         \includegraphics[width=\textwidth]{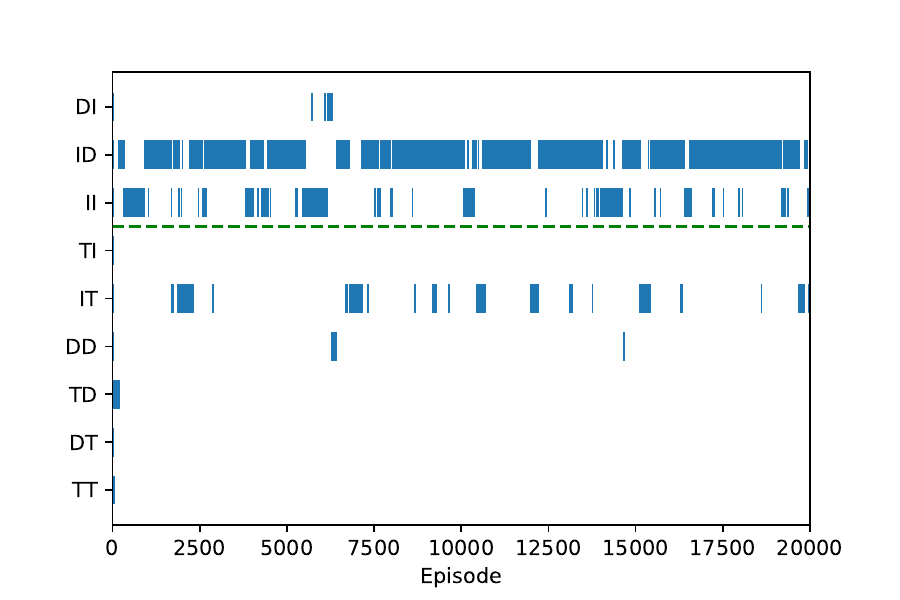}
         \caption{Final policy = II}
         \label{fig:Baseline II}
     \end{subfigure}
     \begin{subfigure}[h]{0.45\textwidth}
         \centering
         \includegraphics[width=\textwidth]{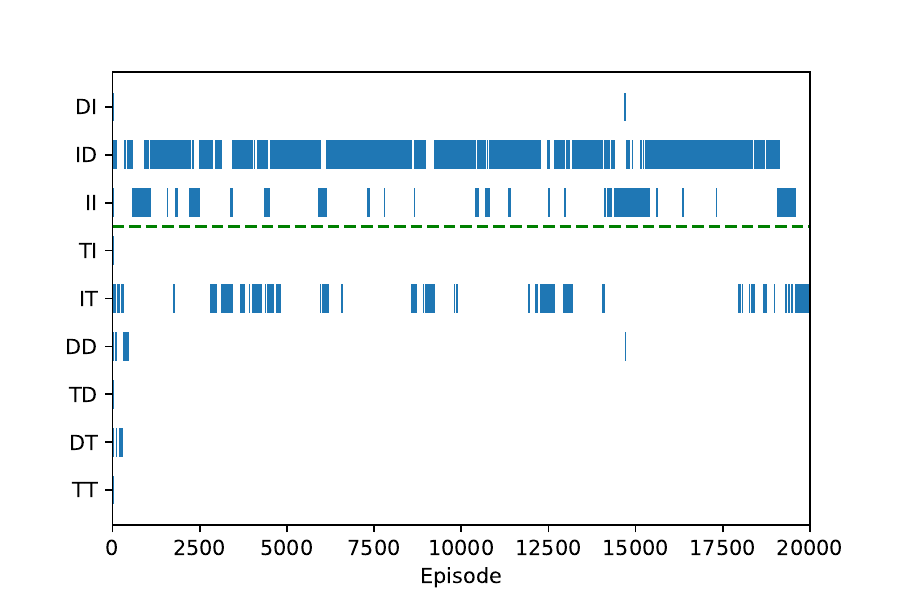}
         \caption{Final policy = IT}
         \label{fig:Baseline IT}
     \end{subfigure}
     \caption[Policy charts for Baseline method]{Policy charts showing the greedy policy produced by the baseline multi-objective Q-learning algorithm (Algorithm \ref{algo:moql-expected}) on the Space Traders environment. Each chart shows the greedy policy identified by the agent at each episode of four different trials, culminating in different final policies. The dashed green line represents the threshold used for TLO, to highlight which policies meet this threshold.}
     \label{fig:Baseline Policy Charts}
\end{figure}

In addition the issue of noisy estimates means that the agent will sometimes settle on another policy, including a policy which does not even meet the success threshold. This is illustrated in Figure \ref{fig:Baseline Policy Charts} which visualises the learning behaviour of the baseline method (Algorithm \ref{algo:moql-expected}) during four of the twenty trials, selected so as to include one example of each of the four different final policies learned by this algorithm, as listed in Table \ref{tab:Space-Traders-Baseline}. Each sub-part of the figure illustrates a single run of the baseline algorithm. For each episode, the policy which the agent believed to be optimal at that stage of its learning is indicated by a blue bar. The green dashed line indicates the threshold for the first objective. The policies on the vertical axis are sorted to indicate that only the DI, ID and II policies meet this constraint. As we can see from all of these policy charts, the agent's behaviour is unstable with frequent changes in its choice of optimal policy. Policy ID is the most frequently selected across 20,000 episodes, which reflects why it is the most frequent final outcome, but in many runs the agent winds up with a different final policy. In particular it can be seen that policies beneath the threshold are regarded as optimal on an intermittent basis, which indicates that the agent's estimate of the value of these policies must be inaccurate.

\begin{figure}[hbt!]
    \centering
    \begin{subfigure}[h]{0.45\textwidth}
         \centering
         \includegraphics[width=\textwidth]{Images/Baseline-ID.pdf}
         \caption{Constant learning rate}
         \label{fig:Original-ID-1}
    \end{subfigure}
    \begin{subfigure}[h]{0.45\textwidth}
         \centering
         \includegraphics[width=\textwidth]{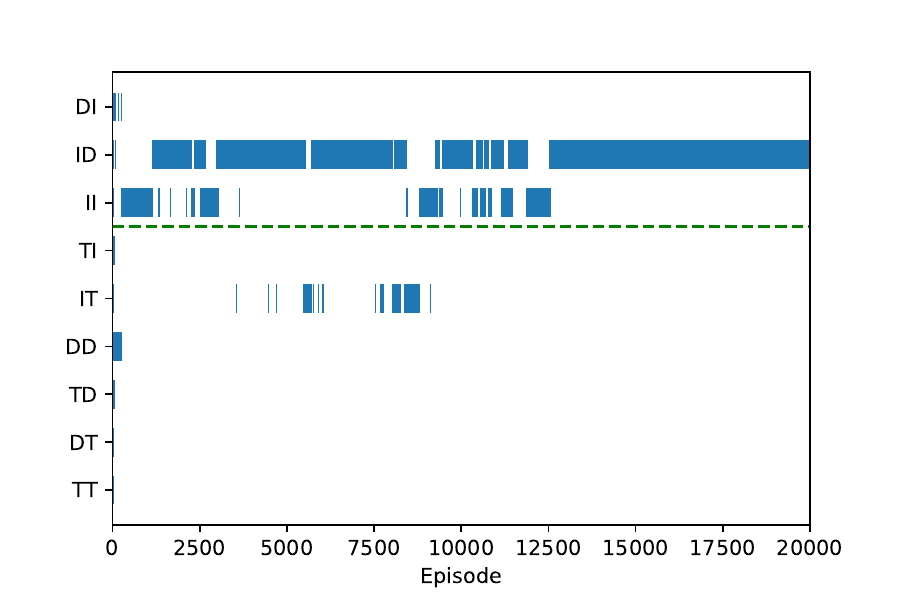}
         \caption{Decayed learning rate}
         \label{fig:d-Original-ID}
    \end{subfigure}
    \caption[Policy charts for baseline method with the decayed learning rate]
    {The Policy chart for baseline method with a decayed learning rate in the Space Traders Environment}
    \label{fig:d-Original-Policy-Chart}
\end{figure}

To highlight the impact of noisy estimates, we ran a further twenty trials during which the learning was linearly decayed from its initial learning rate to zero. All other hyperparameters were the same as in the previous trials of the baseline algorithm. Decaying the learning rate will minimise the impact of the environmental stochasticity on the variation of the agent's Q-values, and should result in increased stability in the choice of greedy policy. Table \ref{tab:Space-Traders-Baseline-d} summarises the final policies learned during these trials, while Figure \ref{fig:d-Original-Policy-Chart} visualises the choice of greedy policy over two representative trials, one with a constant learning rate and one with a decayed learning rate. Both of these trials culminate in the final greedy policy being ID.

\begin{table}[hbt!]
    \centering
    \begin{tabular}{@{}|c|c|c|c|c|@{}}
    \hline
    Policy & DI & ID & II & IT  \\ 
    \hline
    Constant learning rate & 1 & 13 & 4  & 2    \\ 
    \hline
    Decayed learning rate & 0 & 20 & 0  & 0    \\ \hline
    \end{tabular}%
    \caption[Twenty independent runs of baseline multi-objective Q-learning (Algorithm \ref{algo:moql-expected}) with constant and decayed learning rates]{The final greedy policies learned in twenty independent runs of baseline multi-objective Q-learning (Algorithm \ref{algo:moql-expected}) with constant or decayed learning rates for the Space Traders environment}
    \label{tab:Space-Traders-Baseline-d}
\end{table}

As can be seen from Table \ref{tab:Space-Traders-Baseline-d}, the decayed learning rate does indeed result in more stable and consistent learning behaviour, as the agent converges to the same final policy in all twenty trials, compared to the diverse set of final policies evident under a constant learning rate. The policy chart in Figure \ref{fig:d-Original-Policy-Chart} also indicates that gradually decaying the learning rate reduces the influence of the environmental stochasticity. After 15,000 episodes, the agent converges to a single fixed greedy policy until the end of the experiment. So clearly the decayed learning rate helps to eliminate the impact of environmental stochasticity on this agent, allowing it to reliably converge to the same solution. However this also highlights the inability of the baseline method to learn SER-optimal behaviour, as it consistently converges to the sub-optimal policy ID in all twenty trials, never finding the desired DI policy. 

The results in this section of the study reveal two main findings:
\begin{itemize}
  \item The baseline MOQ-learning algorithm fails to learn the SER-optimal policy in the majority of trials (confirming the findings of \cite{VamplewEnvironmental2022}).
  \item The noisy estimates arising from environmental stochasticity lead to instability in the greedy policy learned by MOQ-learning, with variations both within and between trials.
\end{itemize}
\section{Reward Engineering}

In the remainder of the paper we examine various approaches for addressing the inability of the baseline MOQ-learning algorithm to reliably identify the SER-optimal policy for the Space Traders environment. The first approach we consider is to modify the reward structure of Space Traders, while retaining the same environmental dynamics. While the dynamics of state transitions is an intrinsic component of the environment, the reward function is generally specified by a human designer, with the aim of producing the desired behaviour from the agent. Therefore modifying the reward structure is within the designer's control, and a better-designed reward signal may allow for improved performance by the agent. Therefore, the most simple and natural approach is to modify the reward structure first without actually changing the original MOQ-learning algorithm.

\subsection{Modified reward structure and results}
\begin{figure}[hbt!]
    \centering
    \includegraphics[width=12cm]{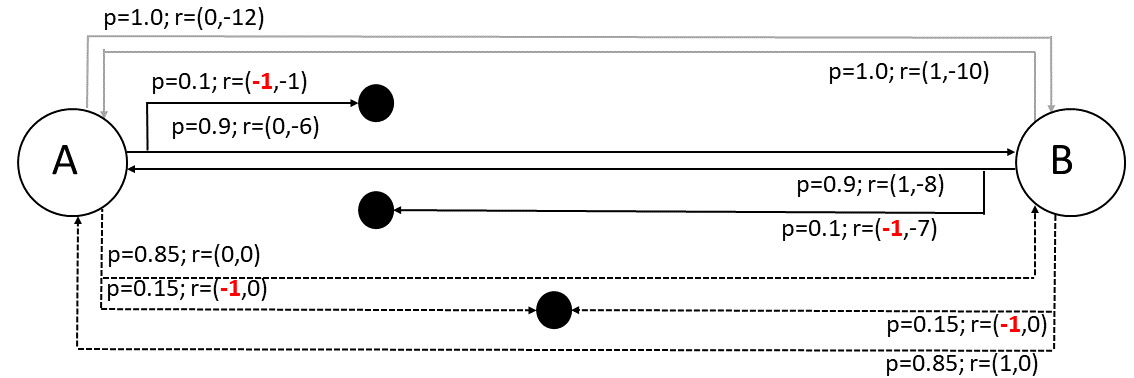}
    \caption[Space Traders MR environment, with a modified reward design. ]
    {The Space Traders MR environment, which has the same state transition dynamics as the original Space Traders but with a modified reward design. The changed rewards have been highlighted in red.}
    \label{fig:ST-V2}
\end{figure}

A version of Space Traders with a modified reward design is shown in Figure \ref{fig:ST-V2} -- we will refer to this as Space Traders MR. The agent will receive a -1 reward for the first objective when visiting one of the terminal states, receive +1 when reaching the goal state, and 0 for other intermediate transitions. The motivation here is to provide additional information to the agent at State A regarding the likelihood of any action leading to a terminal state. As can be seen from the top-half of Table \ref{tab:RewardDesign-Space-Traders-O1}, under the new reward design, the three actions from state A have differing expected immediate rewards for the first objective of 0, -0.1, and -0.15.

As a consequence, the threshold value of the utility function also needs to be updated, because the total rewards for the first element are now ranging from -1 to 1 instead of 0 to 1. Adjusting for this change in range results in a revised threshold equal to $0.88*1+0.12*(-1)=0.76$. All other algorithmic settings remain the same as in Section \ref{sec:baseline}.

\begin{table}[]
    \centering
\begin{tabular}{|l|l|l|l|l|l|}
\hline
State              & Action   & P(success) & \begin{tabular}[c]{@{}l@{}}Reward \\ on\\ success\end{tabular} & \begin{tabular}[c]{@{}l@{}}Reward \\ on\\ failure\end{tabular} & \begin{tabular}[c]{@{}l@{}}Mean\\ reward\end{tabular} \\ \hline
\multirow{3}{*}{A} & Indirect & 1.0        & (0,-12)                                                        & n/a                                                            & (0,-12)                                               \\[5pt] \cline{2-6} 
                   & Direct   & 0.9        & (0, -6)                                                        & (-1, -1)                                                        & (-0.1, -5.5)                                             \\[5pt] \cline{2-6} 
                   & Teleport & 0.85       & (0,0)                                                          & (-1,0)                                                          & (-0.15, 0)                                                \\[5pt] \hline
\multirow{3}{*}{B} & Indirect & 1.0        & (1, -10)                                                       & n/a                                                            & (1, -10)                                              \\[5pt] \cline{2-6} 
                   & Direct   & 0.9        & (1, -8)                                                        & (-1, -7)                                                        & (0.8, -7.9)                                           \\[5pt] \cline{2-6} 
                   & Teleport & 0.85       & (1, 0)                                                         & (-1, 0)                                                         & (0.7, 0)                                             \\[5pt] \hline
\end{tabular}
    \caption[The probability of success and reward values for each state-action pair in Space Traders MR]
    {The probability of success and reward values for each state-action pair in  Space Traders MR}
    \label{tab:RewardDesign-Space-Traders-O1}
\end{table}

\begin{table}[hbt!]
    \centering
    \begin{tabular}{@{}|c|c|c|c|c|c|c|@{}}
    \hline
    Policy & DI & ID & II & IT & TI & DD\\ \hline
    Baseline & 1 & 13 & 4  & 2  & 0 & 0\\ \hline
    Reward Design - constant learning rate & 10 & 5 & 1 & 1 & 2 & 1\\ \hline
    Reward Design - decayed learning rate& 20 & 0 & 0 & 0 & 0 & 0\\ \hline
    \end{tabular}%
    \caption[Twenty independent runs of the Algorithm \ref{algo:moql-expected}]{The final greedy policies learned in twenty independent runs of the Algorithm \ref{algo:moql-expected} for Space Traders MR environment, compared to the original Space Traders environment.}
    \label{tab:Space-Traders-RewardDesign}
\end{table}

\begin{figure}[hbt!]
     \centering
     \begin{subfigure}[h]{0.45\textwidth}
         \centering
         \includegraphics[width=\textwidth]{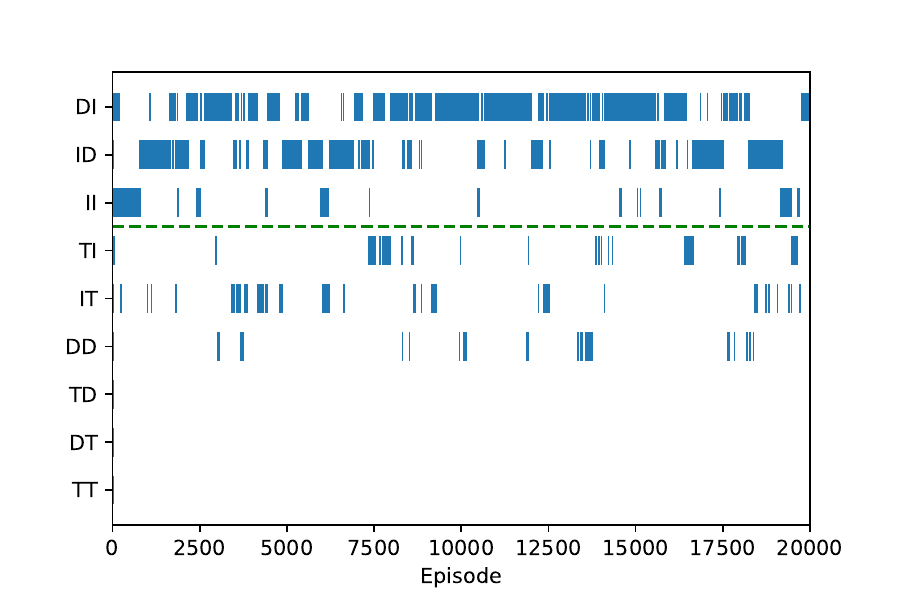}
         \caption{Policy DI}
         \label{fig:RewardDesign DI}
     \end{subfigure}
     \begin{subfigure}[h]{0.45\textwidth}
         \centering
         \includegraphics[width=\textwidth]{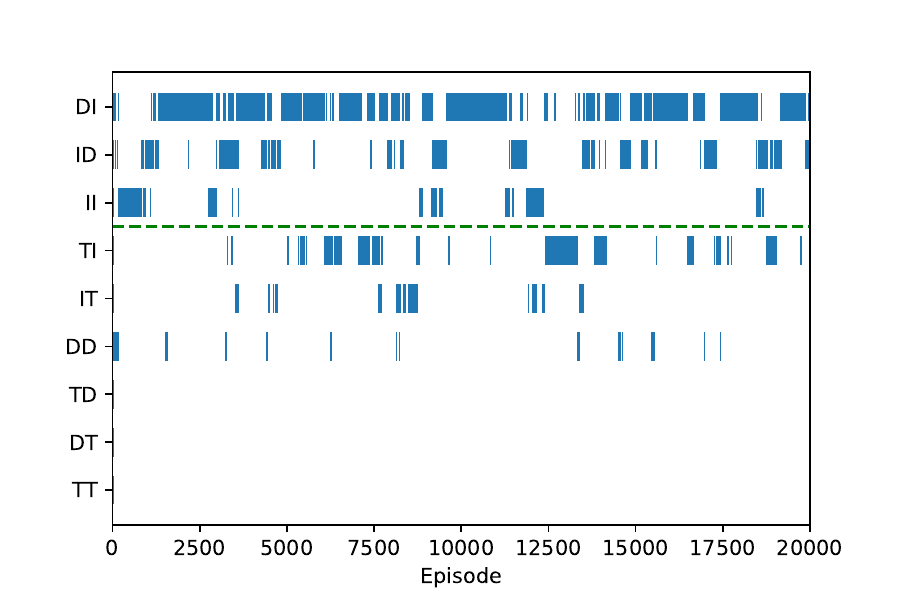}
         \caption{Policy ID}
         \label{fig:RewardDesign ID}
     \end{subfigure}
     \begin{subfigure}[h]{0.45\textwidth}
         \centering
         \includegraphics[width=\textwidth]{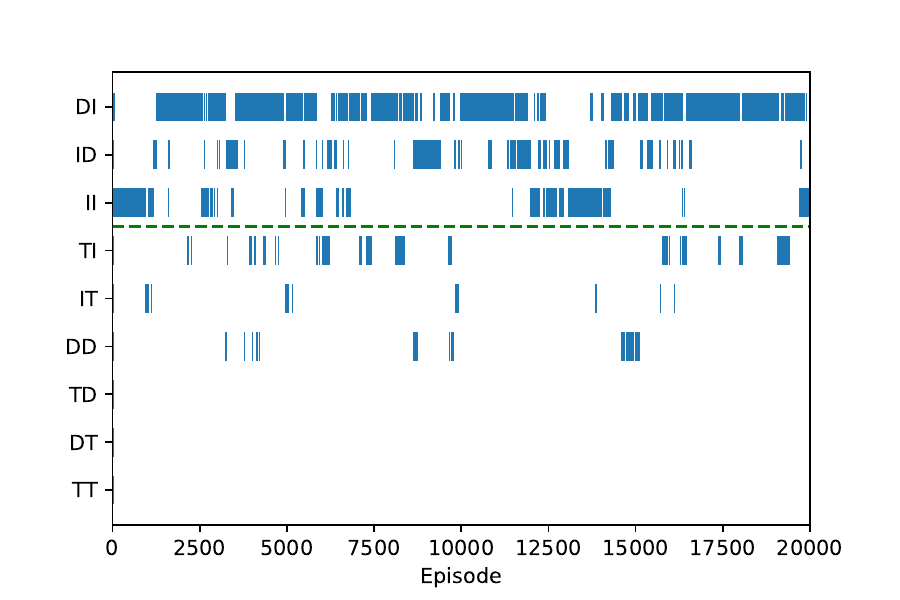}
         \caption{Policy II}
         \label{fig:RewardDesign II}
     \end{subfigure}
     \begin{subfigure}[h]{0.45\textwidth}
         \centering
         \includegraphics[width=\textwidth]{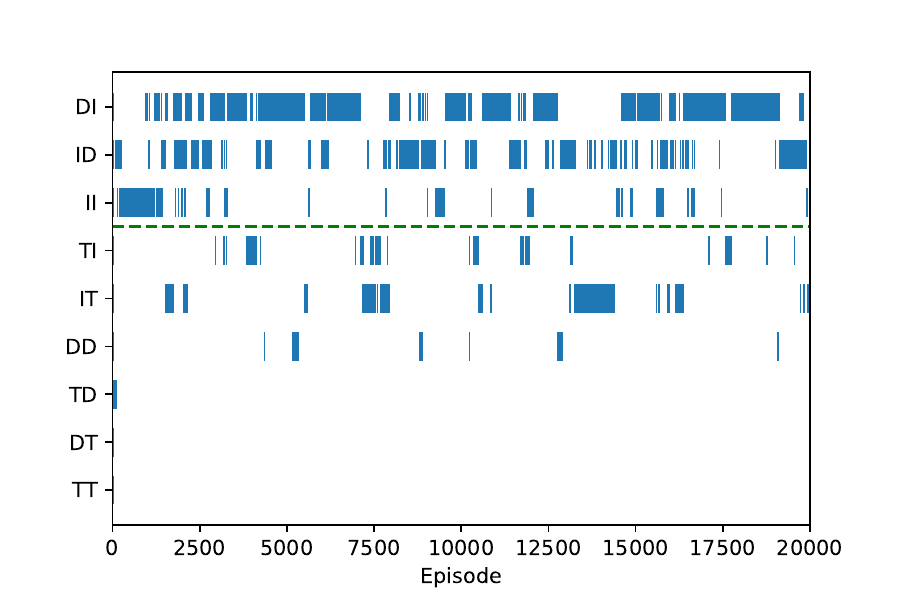}
         \caption{Policy IT}
         \label{fig:RewardDesign IT}
     \end{subfigure}
          \begin{subfigure}[h]{0.45\textwidth}
         \centering
         \includegraphics[width=\textwidth]{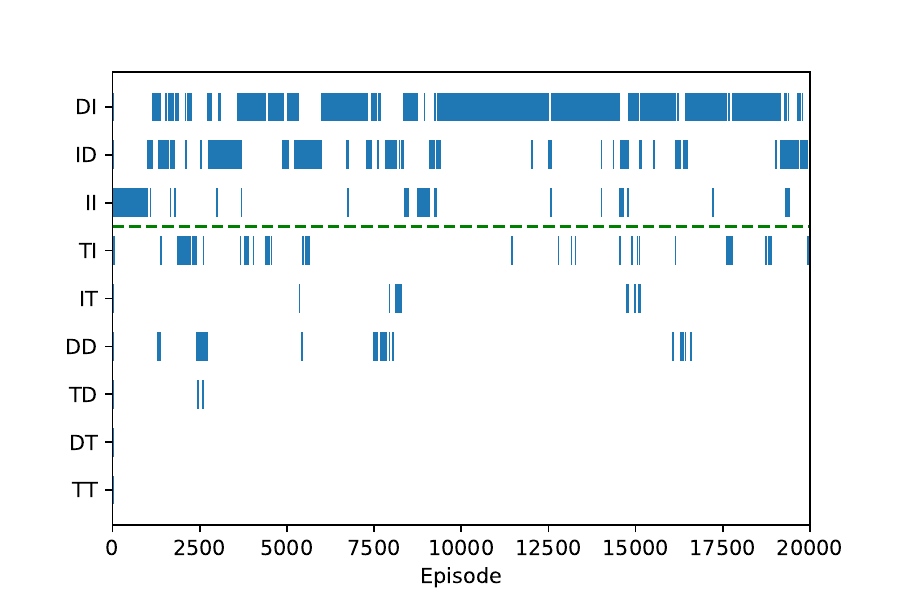}
         \caption{Policy TI}
         \label{fig:RewardDesign TI}
     \end{subfigure}
         \begin{subfigure}[h]{0.45\textwidth}
         \centering
         \includegraphics[width=\textwidth]{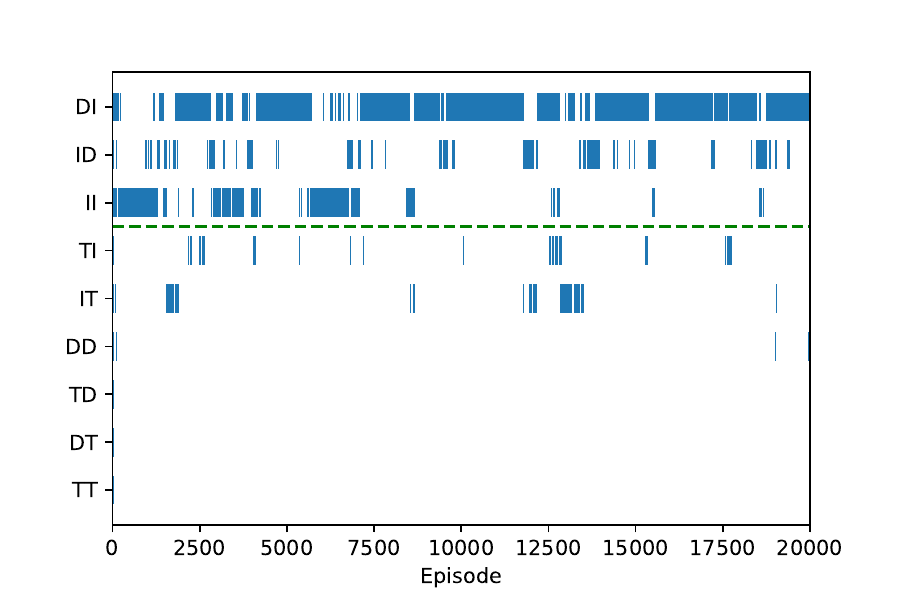}
         \caption{Policy DD}
         \label{fig:RewardDesign DD}
     \end{subfigure}
     \begin{subfigure}[h]{0.45\textwidth}
         \centering
         \includegraphics[width=\textwidth]{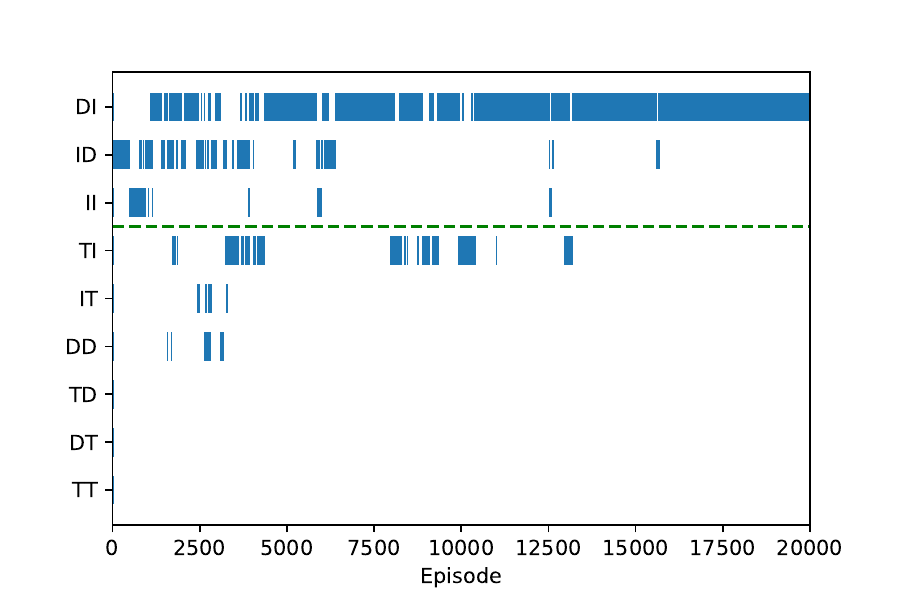}
         \caption{Policy DI - decayed learning rate}
         \label{fig:RewardDesign DD}
     \end{subfigure}
     \caption[Policy charts for MOQ-learning on the Space Traders MR environment]{Policy charts for MOQ-learning on the Space Traders MR environment -- each chart illustrates a sample run culminating in a different final policy. Charts (a)-(f) are from runs using a constant learning rate, while chart (g) is from a run using a decayed learning rate.}
     \label{fig:Reward Design Policy Charts}
\end{figure}

As we can see from Table \ref{tab:Space-Traders-RewardDesign}, the most common outcome (10/20 runs) is the desired DI policy. This is a substantial improvement over the single occurrence of this policy under the original reward design. But on the another hand, the ID policy (5 repetitions) is the second most common outcome, and other policies also occur in some runs. From Figure \ref{fig:Reward Design Policy Charts}, most of time across 20,000 episodes, the agent prefers policy DI which is our desired optimal policy. However the intermittent identification of the other policies as optimal means that overall this approach still only yields the correct policy 50\% of the time. It is also worth noting that this reward structure results in an even greater variety of sub-optimal solutions being found compared to the original reward design.

To demonstrate that the trials in which the agent converges to the non-optimal policy for Space Traders MR are due to noisy estimates, we conducted a further set of experiments applying an agent with a decayed learning rate to this environment. As shown in the final row of Table \ref{tab:Space-Traders-RewardDesign}, this agent converges to the optimal DI policy in all 20 trials. The policy charts in Figure \ref{fig:Reward Design Policy Charts} highlight the difference between learning with a constant and decayed learning rate. While the runs shown in (a) and (g) both settle on DI as their final policy, the latter run (using a decayed learning rate) is far more stable than the former (which uses a constant learning rate, and which still selects policies other than DI as optimal even late in the training process). This confirms the earlier observations from the Baseline experiments that noisy estimates can have a significant impact on the stability and accuracy of learning.

\subsection{Space Traders with modified state structure}

The results in the previous section show that modifying the reward structure to explicitly provide a negative reward component on transitions to fatal terminal states allows for improved performance from the MOQ-learning algorithm on the SpaceTraders environment. However in order for this approach to be useful, we need to confirm that similar reward structures which capture relevant information about transitions to terminal states are possible regardless of the structure of the environment and its state dynamics. 

\begin{figure}[hbt!]
    \centering
    \includegraphics[width=12cm]{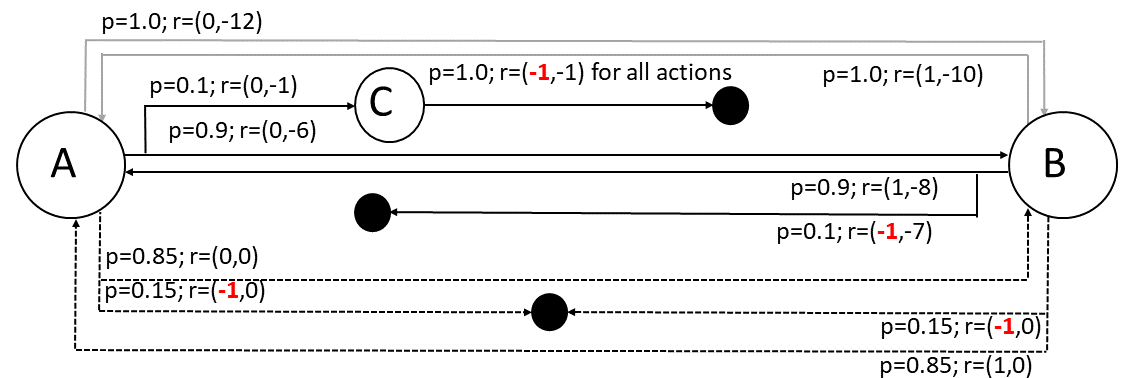}
    \caption[The new variant of reward design for Space Traders Environment]
    {The Space Traders 3-State environment which adds an additional state C to the Space Traders  MOMDP with a more complex state structure. All the changes have been highlighted in red color}
    \label{fig:ST-V3}
\end{figure}

To investigate this, we introduce a variant of the Space Traders environment with an additional state as shown in Figure \ref{fig:ST-V3} -- this will be referred to as Space Traders 3-State (3ST). It includes a new state C reached when the agent selects the direct action at state A. This introduces a delay between the selection of the direct action at A, and the ultimate reaching of the terminal state (and consequent negative reward for the first objective).  

The empirical results from twenty trials show that the desired optimal policy (DI) was not converged to in practice for the Space Traders 3-State environment. A closer examination of the behaviour of the agent on Space Traders MR shows that at state B the agent will have different accumulated expected reward for the first objective depending on which action was chosen at State A. For example, if the agent selects the Direct action and successfully reaches state B, the ideal accumulated expected reward for the first objective will be $0.9*0+0.1*(-1)=-0.1$ when the action values are learned with sufficient accuracy. This will lead the agent to select the indirect action in state B. But in the Space Traders 3-State variant environment, the accumulated expected reward will be zero when the agent reaches state B by taking direct action (as was also the case for the original Space Traders MOMDP). As a result, the agent converges to the sub-optimal policy ID in practice again.

\begin{table}[hbt!]
    \centering
    \begin{tabular}{@{}|c|c|c|c|c|c|c|c|@{}}
    \hline
    Policy & DI & ID & II & IT & TI & DD & TD\\ \hline
    Reward Design & 10 & 5 & 1 & 1 & 2 & 1 & 0\\ \hline
    Extra State & 0 & 14 & 2 & 1 & 2 & 0 & 1\\ \hline
    \end{tabular}%
    \caption[Twenty independent runs of the Algorithm \ref{algo:moql-expected}]{The final greedy policies learned in twenty independent runs of the Algorithm \ref{algo:moql-expected} for Space Traders 3-State environment, compared to the Space Traders MR environment.}
    \label{tab:Space-Traders-RewardDesign-v}
\end{table}

This example illustrates that while it may be possible in some cases to encourage SER-optimal behaviour via a careful designing of rewards, more generally the structure of the environment may make it difficult or impossible to identify a suitable reward design. The modified reward used for Space Traders MR was based on a simple principle of providing a -1 reward for the success objective on any transitions to a terminal state. For this particular environment structure, this reward signal essentially captures the required information such as the transition probabilities within the accumulated expected reward for the first objective. However, as shown by the Space Traders 3-State variant, this reward design principle is not sufficient in general. Therefore, relying on the reward designer being able to create a suitable reward structure is insufficient to provide a general means to address issues in stochastic environments under SER criteria.

\section{Incorporation of Global Statistics}

As identified previously, the main issue for applying MOQ-learning algorithm to stochastic environment is that the action selection at given state is purely based on local information (the Q values for the current state) and current episode information (accumulated expected reward)\cite{VamplewEnvironmental2022}. This is the same issue previously identified for multi-objective planning algorithms by \cite{bryceProbabilisticPlanningMultiobjective2007}. However, in order to maximise the expected utility over multiple episodes (SER criteria) the agent must also consider the expected return on other episodes where the current state is not reached. In other words, the agent must also have some level of knowledge about global statistics in order to maximise \acrfull{SER}. Therefore the second approach we examine is to include extra global information within the MOQ-learning algorithm.

\subsection{Multi-objective Stochastic State Q-learning (MOSS)}
To support this idea, Multi-objective Stochastic State Q-learning (MOSS)(Algorithm \ref{algo:mossql}) is introduced\footnote{This algorithm was named by the second author in honour of IT pioneer Maurice Moss.}. Here are the changes compared with previous MOQ-learning (Algorithm \ref{algo:moql-expected})
\begin{itemize}
  \item The agent maintains two pieces of global information: the total number of episodes experienced ($v_\pi$), and an estimate of the average per-episode return ($E_\pi$).
  \item For every state, the agent maintains a counter of episodes in which this state was visited at least once ($v(s)$), and the estimated average return in those episodes ($E(s)$).
  \item When selecting an action, the agent uses those values to estimate the average return in episodes where the current state is not visited. This value is then combined with the estimated accumulated rewards $P(s)$ and Q-value $Q(s)$ to estimate the return for each action, taking into account all episodes (both the episodes in which this state is visited, and those in which it is not visited). Action selection is then based on this holistic measure of the value for each action, which should make the action selection more compatible with the goal of finding the SER-optimal policy.
\end{itemize}

\begin{algorithm}[hbt!]
  \caption{The multi-objective stochastic state Q($\lambda$) algorithm (MOSS Q-learning). Highlighted text identifies the changes and extensions introduced relative to multi-objective Q($\lambda$) as previously described in Algorithm \ref{algo:moql-expected}.}
  \label{algo:mossql}
  \begin{algorithmic}[1]
    \Statex input: learning rate $\alpha$, discounting term $\gamma$, eligibility trace decay term $\lambda$, number of objectives $n$, action-selection function $f$ and any associated parameters
    \For {all states $s$, actions $a$ and objectives $o$}
    	\State initialise $Q_o(s,a)$
    	\textcolor{red}{
    	    \State initialise $P_o(s)$ \Comment{expected cumulative reward when $s$ is reached}
    	    \State initialise $v(s)=0$ \Comment{count of visits to s}
    	}
    \EndFor
    \State \textcolor{red}{initialise $E_\pi$ \Comment{estimated return over all episodes}
    }
    \State \textcolor{red}{initialise $v_\pi=0$ \Comment{count of all episodes}}
   \For {each episode} 
            \State \textcolor{red}{$v_\pi = v_\pi + 1$
            \Comment{increment episode counter}}
        \For {all states $s$ and actions $a$}
    		\State $e(s,a)$=0; 
    		\textcolor{red}{ $b(s)=0$ \Comment{binary flag - was $s$ visited in this episode?}
        }
        \EndFor
        \State sums of prior rewards $P_o$ = 0, for all $o$ in 1..$n$
    	\State observe initial state $s_t$
    	\textcolor{red}{
    	    \\ \Comment{call Algorithm \ref{algo:update-stats1} to update stats and create augmented state and utility vector}
    	    \State $s^A_t$, $U(s^A_t)$ = update-statistics($s_t$,$P$)
            \State select $a_t$ from an exploratory policy derived using $f(U(s^A_t))$
        }\strut
        \For {each step of the episode}
 			\State execute $a_t$, observe $s_{t+1}$ and reward $R_t$
 			\State $P = P + R_t$
    	    \textcolor{red}{
    	        \State $s^A_{t+1}$, $U(s^A_{t+1})$ = update-statistics($s_{t+1}$,$P$)
            } 			
 			\State select $a^*$ from a greedy policy derived using  $f(U(s^A_{t+1}))$
  			\State select $a^\prime$ from an exploratory policy derived using $f(U(s^A_{t+1}))$
            \State $\delta = R_t + \gamma Q(s^A_{t+1},a^*) - Q(s^A_t,a_t)$
            \State $e(s^A_t,a_t)$ = 1
            \For {each augmented state $s^A$ and action $a$}
            	\State $Q(s^A,a) = Q(s^A,a) + \alpha\delta e(s^A,a)$
                 \If {$a^\prime = a^*$} 
                    \State $e(s^A,a) = \gamma \lambda e(s,a)$
                 \Else
                    \State $e(s^A,a) = 0$                 
                 \EndIf
            \EndFor
            \State $s^A_t = s^A_{t+1}, a_t = a^\prime$
    	\EndFor
    	\textcolor{red}{
    	    \State $E_\pi = E_\pi + \alpha(P-E_\pi)$ \Comment update estimates of per-episode return
            \For {all states with $b(s)\neq0$}
              \State $E(s) = E(s) + \alpha(P-E(s))$
            \EndFor
        }    	
    \EndFor
  \end{algorithmic}
\end{algorithm}

\begin{algorithm}[hbt!]
  \caption{The update-statistics helper algorithm for MOSS Q-learning (Algorithm \ref{algo:mossql}). Given a particular state $s$ it updates the global variables which store statistics related to $s$. It will then return an augmented state formed from the concatenation of $s$ with the estimated mean accumulated reward when $s$ is reached, and a utility vector $U$ which estimates the mean vector return over all episodes for each action available in $s$}
  \label{algo:update-stats1}
  \begin{algorithmic}[1]
    \Statex input: state $s$, accumulated rewards in the current episode $P$
    \If {$b(s)=0$} \Comment{first visit to $s$ in this episode}
        \State $v(s) = v(s) + 1$ \Comment{increment count of visits to $s$}
        \State $b(s)=1$ \Comment{set flag so duplicate visits within an episode are not counted}
    \EndIf
    \State $P(s) = P(s) + \alpha (P-P(s))$
    \State $s^A$ = $(s,P(s))$ \Comment{augmented state}
    \State $p(s) = v(s)/v_\pi$    \Comment{estimated probability of visiting $s$ in any episode}
    \If {p(s)=1} \Comment{treat states which are always visited as a special case}
        \For {each action $a$}
            \State $U(a) = P(s)+Q(s^A,a))$ 
        \EndFor
    \Else
        \State $E_{\not{s}}=(E_\pi - p(s)E_s)/(1-p(s)$ \Comment estimated return in episodes where $s$ is not visited
        \State \Comment{calculate estimated value over all episodes, assuming a is executed in $s^A$}
        \For {each action $a$}
            \State $U(a) = p(s)(P(s)+Q(s^A,a)) + (1-p(s))E_{\not{s}}$ 
        \EndFor
    \EndIf
    \State return $s^A$, $U$
  \end{algorithmic}
\end{algorithm}

\subsection{MOSS Q-learning Results}
\begin{figure}[hbt!]
     \centering
     \begin{subfigure}[h]{0.45\textwidth}
         \centering
         \includegraphics[width=\textwidth]{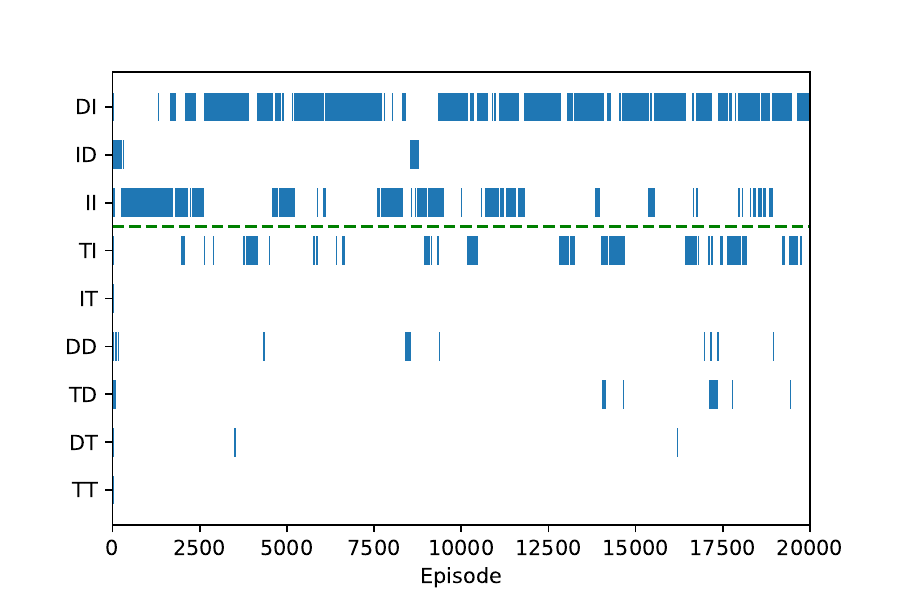}
         \caption{Policy DI}
         \label{fig:MOSS DI}
     \end{subfigure}
     \begin{subfigure}[h]{0.45\textwidth}
         \centering
         \includegraphics[width=\textwidth]{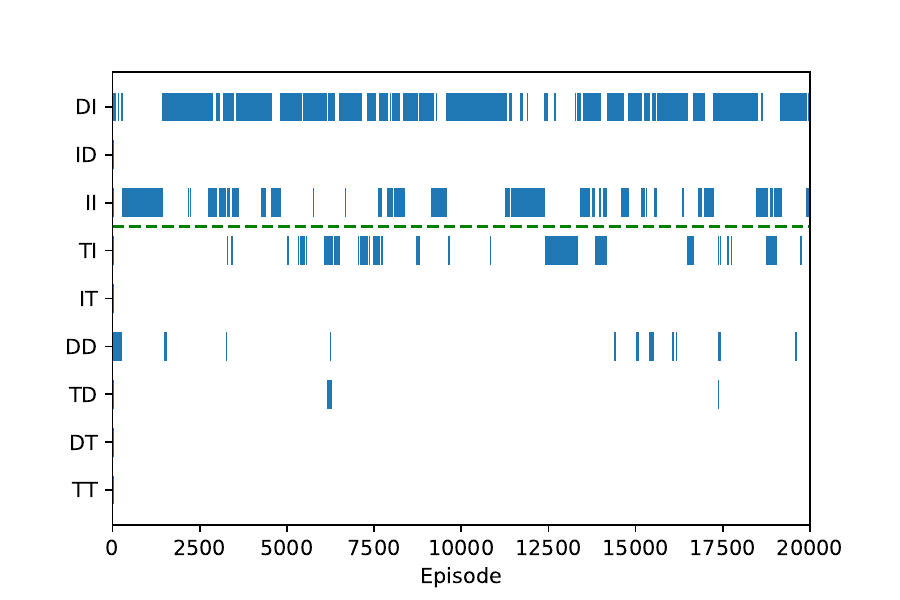}
         \caption{Policy II}
         \label{fig:MOSS II}
     \end{subfigure}
     \begin{subfigure}[h]{0.45\textwidth}
         \centering
         \includegraphics[width=\textwidth]{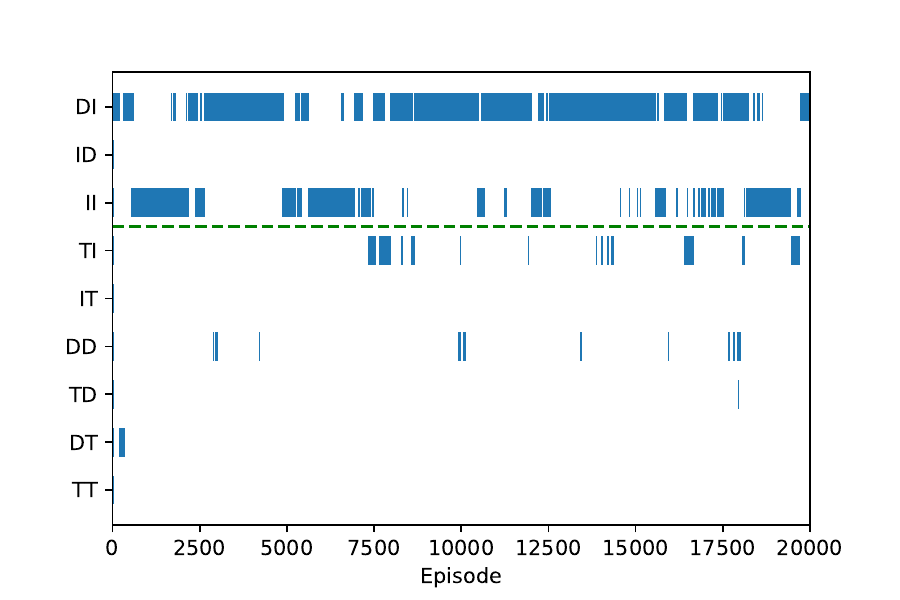}
         \caption{Policy TI}
         \label{fig:MOSS TI}
     \end{subfigure}
     \caption[Policy charts for Single-Phase MOSS]{3 policy charts for Single-Phase MOSS Algorithm}
     \label{fig:MOSS Policy Charts}
\end{figure}
\begin{table}[hbt!]
    \centering
    \begin{tabular}{@{}|c|c|c|c|c|c|c|@{}}
    \hline
    Policy & DI & ID & II & IT & TI & DD\\ \hline
    Baseline & 1 & 13 & 4  & 2  & 0 & 0\\ \hline
    MOSS & 15 & 0 & 0 & 3 & 2 & 0\\ \hline
    \end{tabular}%
    \caption[Twenty independent runs of the Algorithm \ref{algo:mossql}]{The final greedy policies learned in twenty independent runs of the MOSS Q-learning algorithm for the Space Traders environment}
    \label{tab:Space-Traders-MOSS}
\end{table}
As we can see from Table \ref{tab:Space-Traders-MOSS}, the most common result (15/20 runs)
is the DI policy, which is the desired optimal policy, but the IT policy (3 repetitions) and TI policy (2) also occur in some trials. Figure \ref{fig:MOSS Policy Charts} shows that while the agent believes policy DI is the desired optimal policy most of the time, its choice of policy is unstable and occasionally  Policy IT or TI are preferred at the end of training.

\begin{figure}[hbt!]
    \centering
    \includegraphics[width=12cm]{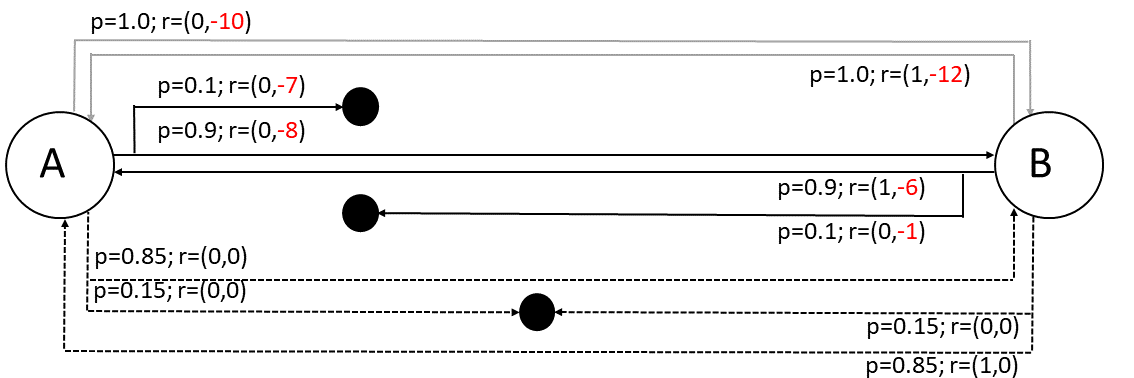}
    \caption[New variant of Space Traders Environment]
    {The Space Traders ID variant environment. All the changes compared with original have been highlighted in red. The changed rewards result in ID being the SER-optimal policy for this environment.}
    \label{fig:ST-V1}
\end{figure}

Based on the results in Table \ref{tab:Space-Traders-MOSS}, the MOSS algorithm clearly outperforms the baseline method in original Space Traders problem, although it still fails to find the optimal policy in many trials. In order to further test this MOSS algorithm we introduce a further variant of the Space Traders Problem (Space Traders ID) as shown in Figure \ref{fig:ST-V1}.

\begin{table}[]
    \centering
\begin{tabular}{|l|l|l|l|l|l|}
\hline
State              & Action   & P(success) & \begin{tabular}[c]{@{}l@{}}Reward \\ on\\ success\end{tabular} & \begin{tabular}[c]{@{}l@{}}Reward \\ on\\ failure\end{tabular} & \begin{tabular}[c]{@{}l@{}}Mean\\ reward\end{tabular} \\ \hline
\multirow{3}{*}{A} & Indirect & 1.0        & (0,-10)                                                        & n/a                                                            & (0,-10)                                               \\[5pt] \cline{2-6} 
                   & Direct   & 0.9        & (0, -8)                                                        & (0, -7)                                                        & (0, -7.9)                                             \\[5pt] \cline{2-6} 
                   & Teleport & 0.85       & (0,0)                                                          & (0,0)                                                          & (0, 0)                                                \\[5pt] \hline
\multirow{3}{*}{B} & Indirect & 1.0        & (1, -12)                                                       & n/a                                                            & (1, -12)                                              \\[5pt] \cline{2-6} 
                   & Direct   & 0.9        & (1, -6)                                                        & (0, -1)                                                        & (0.9, -5.5)                                           \\[5pt] \cline{2-6} 
                   & Teleport & 0.85       & (1, 0)                                                         & (0, 0)                                                         & (0.85, 0)                                             \\[5pt] \hline
\end{tabular}
        \caption[The State Action pair in the new variant Space Traders Environment]
    {The probability of success and reward values for each state-action pair in the new variant Space Traders ID environment}
    \label{tab:Space-Traders-MOSS-O1}
\end{table}

\begin{table}[hbt!]
    \centering
    \begin{tabular}{@{}|c|c|c|c|@{}}
    \hline
    Policy identifier & Action in state A & Action in state B & Mean Reward     \\ \hline
    II                & Indirect          & Indirect          & (1, -22)        \\ \hline
    ID                & Indirect          & Direct            & (0.9, -15.5)    \\ \hline
    IT                & Indirect          & Teleport          & (0.85, -10)     \\ \hline
    DI                & Direct            & Indirect          & (0.9, -18.7)    \\ \hline
    DD                & Direct            & Direct            & (0.81, -12.85)  \\ \hline
    DT                & Direct            & Teleport          & (0.765, -7.9)   \\ \hline
    TI                & Teleport          & Indirect          & (0.85, -10.2)    \\ \hline
    TD                & Teleport          & Direct            & (0.765, -4.675) \\ \hline
    TT                & Teleport          & Teleport          & (0.7225, 0)     \\ \hline
    \end{tabular}
    \caption[Mean return for nine deterministic policies in new variant Space Traders Environment]{Nine available deterministic policies mean return for Space Traders ID environment}
    \label{tab:Space-Traders-MOSS-O2}
\end{table}
\begin{figure}[hbt!]
     \centering
     \begin{subfigure}[h]{0.45\textwidth}
         \centering
         \includegraphics[width=\textwidth]{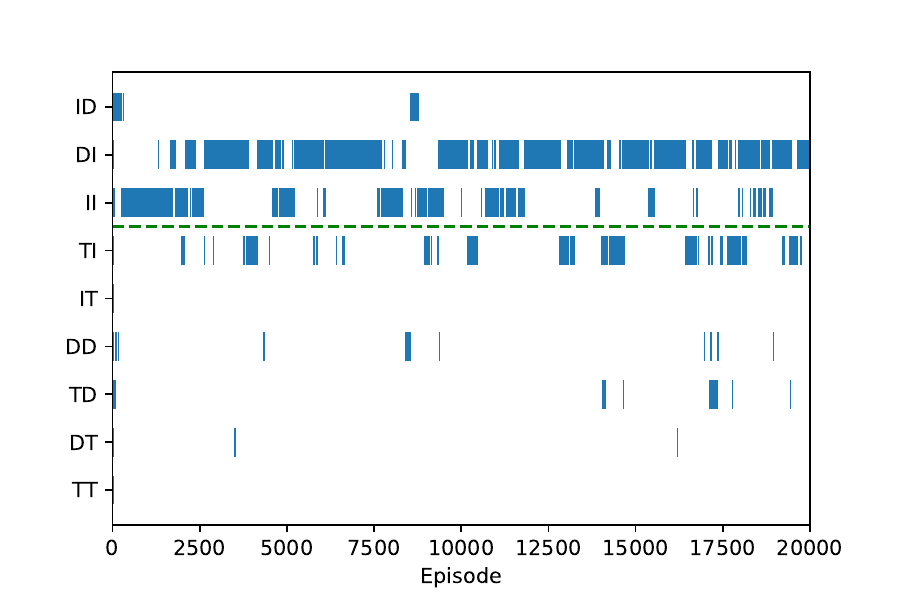}
         \caption{Policy DI}
         \label{fig:MOSS-v DI}
     \end{subfigure}
     \begin{subfigure}[h]{0.45\textwidth}
         \centering
         \includegraphics[width=\textwidth]{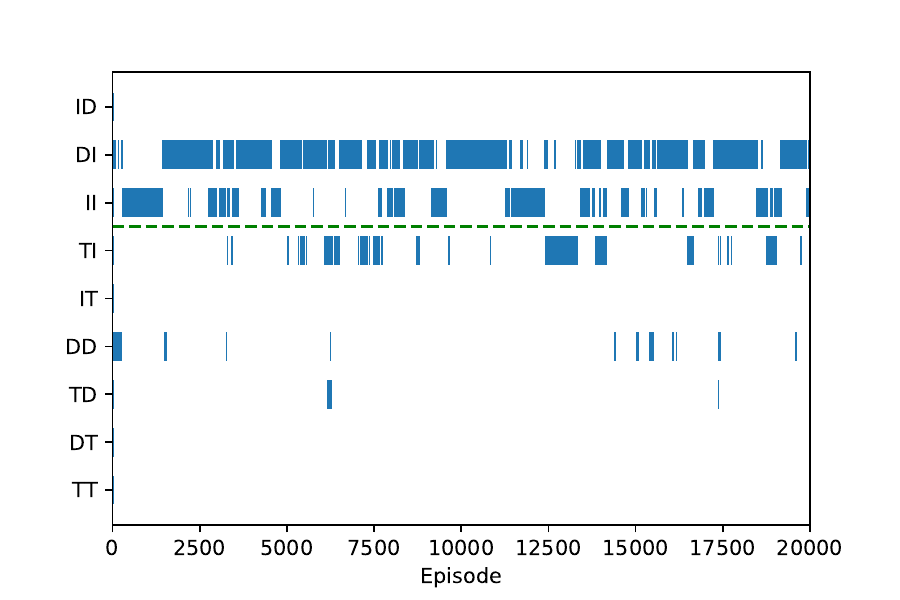}
         \caption{Policy II}
         \label{fig:MOSS-v II}
     \end{subfigure}
     \begin{subfigure}[h]{0.45\textwidth}
         \centering
         \includegraphics[width=\textwidth]{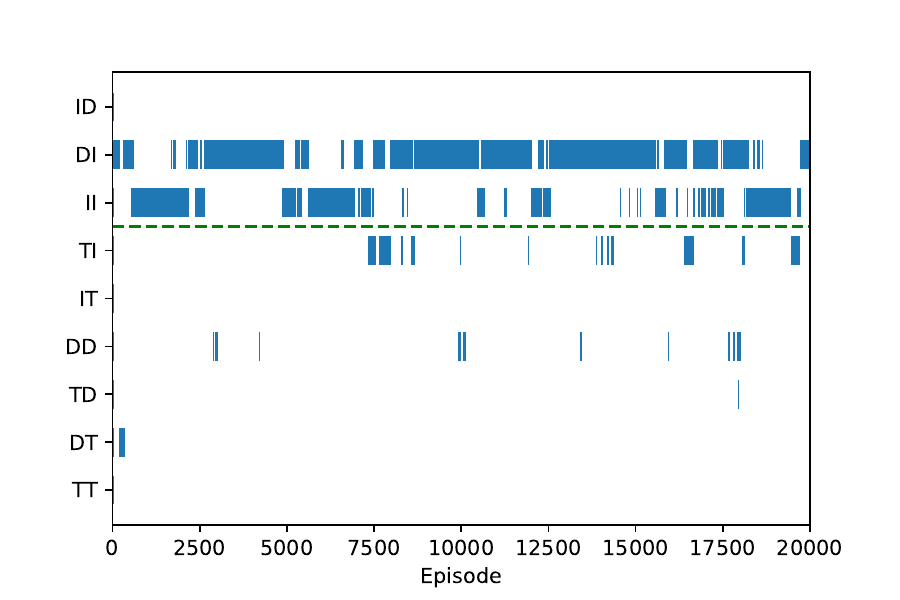}
         \caption{Policy TI}
         \label{fig:MOSS-v TI}
     \end{subfigure}
     \caption[Policy charts for MOSS Q-learning in the Space Traders ID MOMDP.]{Three representative policy charts for MOSS Q-learning in the Space Traders ID MOMDP.}
     \label{fig:MOSS-v Policy Charts}
\end{figure}
\begin{table}[hbt!]
    \centering
    \begin{tabular}{@{}|c|c|c|c|c|c|c|@{}}
    \hline
    Policy & DI & ID & II & IT & TI & DD\\ \hline
    Original & \textcolor{red}{15} & 0 & 0 & 3 & 2 & 0\\ \hline
    New variant & 15 & \textcolor{red}{0} & 0 & 3 & 2 & 0\\ \hline
    \end{tabular}%
    \caption[Twenty independent runs of the Algorithm \ref{algo:mossql}]{The final greedy policies learned in twenty independent runs of MOSS Q-learning for the Space Traders and Space Traders ID environments. The red color indicates the desired optimal policy.}
    \label{tab:Space-Traders-MOSS-v}
\end{table}

All the changes compared with original one have been highlighted in red. The main difference is that the time penalty for each action has been swapped from state A to state B. The new probability of success and reward values for each state-action pair in the new variant Space Traders are shown in Table \ref{tab:Space-Traders-MOSS-O1}. The only difference between policies DI and ID in the original Space Traders Problem is the second objective - the time penalty. Therefore in this new variant of Space Traders problem, policy ID has become the desired SER-optimal policy as we can see from Table \ref{tab:Space-Traders-MOSS-O2}.\newline

The results in Table \ref{tab:Space-Traders-MOSS-v} show that the desired optimal policy (ID) was not converged to in practice, as the most common result (15/20 runs) is still policy DI. As shown by the policy charts (Figure \ref{fig:MOSS-v Policy Charts}), the
agent prefers policy DI most of the time across 20,000 episodes, just as it did in the original Space Trader problem. A closer examination of the MOSS algorithm \ref{algo:mossql} reveals that the estimated values on which action-selection is based ($s_t$, $P(s_t)$, $p(s_t)$ and $E_{\not{s_{t+1}}}$) should be based only on the trajectories produced during execution of the greedy policy, whereas in the current algorithm they are derived from all trajectories. As a result, the value of $p(s_t)$ \footnote{The estimated probability of visiting state s in any episode.} is below 1 because of exploratory actions. In turn the $U(a)$ values at state B for the direct and teleport actions are below threshold for first objective. So it can already be seen that this agent will not converge to the desired policy ID.\footnote{We speculated that this might be addressed using a two-phase variant of MOSS which had separate learning and global statistics gathering phases, with the latter based strictly on the agent's current greedy policy. However this failed to overcome the issues reported here, and so for reasons of space and clarity we have omitted that algorithm from this paper. Full details are available in \cite{ding2022addressing}.}

We also investigated whether noisy estimates may be the cause of the runs in which MOSS fails to converge to the SER-optimal policy for both variations of the Space Traders environments. As before this was achieved by linearly decaying the learning rate from its initial value to zero over the course of learning. 

\begin{figure}[hbt!]
    \centering
    \begin{subfigure}[h]{0.45\textwidth}
         \centering
         \includegraphics[width=\textwidth]{Images/MOSS-DI.pdf}
         \caption{Policy DI with constant learning rate}
         \label{fig:d-MOSS-ID-1}
    \end{subfigure}
     \begin{subfigure}[h]{0.45\textwidth}
         \centering
         \includegraphics[width=\textwidth]{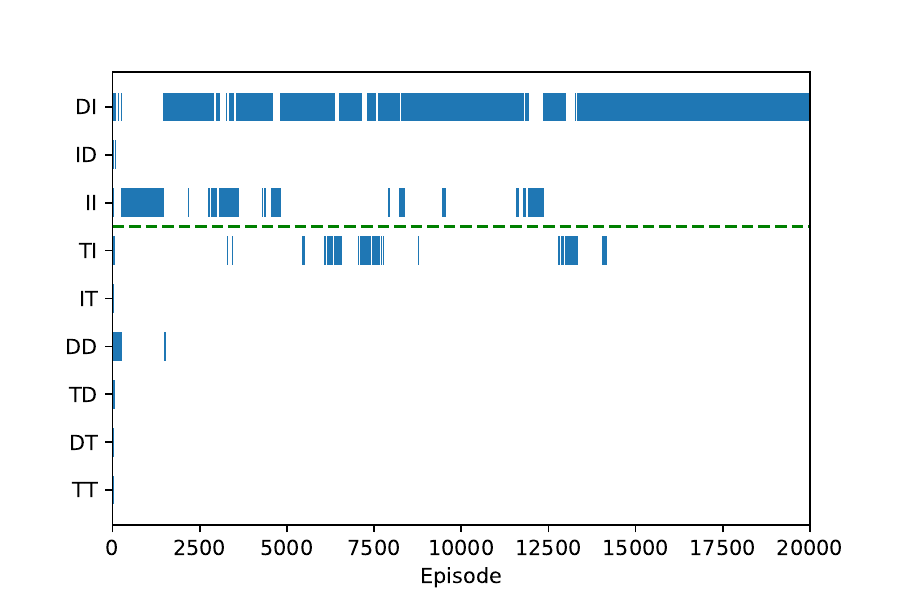}
         \caption{Policy DI with decayed learning rate}
         \label{fig:d-MOSS-ID}
    \end{subfigure}
    \caption[The Policy chart for MOSS algorithm with the decayed learning rate]
    {The Policy chart for the MOSS algorithm with the decayed learning rate in original Space Traders Environment}
    \label{fig:d-MOSS-Policy-Chart}
\end{figure}
        
\begin{table}[hbt!]
    \centering
    \begin{tabular}{@{}|c|c|c|c|c|c|c|@{}}
    \hline
    Environment & Learning rate & DI & ID & II & IT & TI \\ \hline
    Space Traders&Constant &\textcolor{red}{15} & 0 & 0 & 3 & 2 \\ \hline
    Space Traders&Decayed&\textcolor{red}{20} & 0 & 0 & 0 & 0 \\ \hline
    Space Traders ID &Constant & 15 & \textcolor{red}{0} & 0 & 3 & 2 \\ \hline
    Space Traders ID &Decayed & 20 & \textcolor{red}{0} & 0 & 0 & 0 \\ \hline
    \end{tabular}%
    \caption[The final greedy policies learned in twenty independent runs of the MOSS algorithm with either a constant or decayed learning rate for both the Space Traders and Space Traders ID environments.]{The final greedy policies learned in twenty independent runs of the MOSS algorithm with either a constant or decayed learning rate for both the Space Traders and Space Traders ID environments. Red text highlights the SER-optimal policy for each environment.}
    \label{tab:Space-Traders-MOSS-v-d}
\end{table}

The policy charts in Figure \ref{fig:d-MOSS-Policy-Chart} show that the decaying learning rate does improve the stability and consistency of learning. MOSS with a decayed learning rate successfully stabilises on the desired optimal policy DI around 15,000 episodes in Figure \ref{fig:d-MOSS-Policy-Chart}, compared with the policy chart on the left where agent is still struggling to stabilise the final policy before the end of experiment. The results in Table \ref{tab:Space-Traders-MOSS-v-d} show that, for the original Space Traders environment, the combination of the MOSS algorithm and a decayed learning rate does reliably converge to the correct SER-optimal policy DI. However when we apply them to Space Traders ID, the agent fails to learn the desired optimal policy ID. While the decayed learning rate improves the consistency of the algorithm, for this environment that simply means that it consistently converges to the same incorrect policy. 

Therefore, even if the issue of noisy estimates is succesfully eliminated, the MOSS algorithm is not an adequate solution to the problem of learning SER-optimal policies for stochastic MOMDPs.

\section{Policy Options}

\subsection {Policy options algorithm}
The final approach we examine to address the issue of SER-optimality is based on the concept of options. An option is a temporally-extended action, consisting of a sequence of single-step actions which the agent commits to in advance, as opposed to selecting an action on each time-step \cite{sutton1999between}. Often it is used as a means of accelerating learning by embedding pre-existing human knowledge in the form of options. The agent selects an option and executes the sequence of actions defined by it, while continuing to observe states and rewards on each time-step, and learning the Q-values associated with the fine-grained actions.

The use we make of options here differs from their usual application. The simplicity of the original Space Traders environment (2 states, 3 actions per state) means it is possible to define nine options corresponding to the nine deterministic policies which we know exist for this environment. At the start of each episode the agent selects one of these \emph{policy options} to perform, and precommits to following that policy for the entire episode. Therefore rather than learning state-action values for all states, it is sufficient for the agent to just learn option values for the starting state (State A). Over time the state-option values the agent has learnt at state A should match the mean rewards for each of the nine deterministic policies in Table \ref{tab:Space-Traders-O2}. This policy-options approach is detailed in Algorithm \ref{algo:moql-options}.

Performing action-selection in advance based on the estimated values of each policy eliminates the local decision-making at each state which has been identified as the cause of the issues which methods like multi-objective Q-learning have in learning SER-optimal policies \cite{bryceProbabilisticPlanningMultiobjective2007,VamplewEnvironmental2022}. Clearly such an approach is infeasible for more complex environments, as the number of deterministic policies will equal ${|A|}^{|S|}$ and so grows extremely rapidly as the number of actions and states extends beyond the 3 actions and 2 states which exist in SpaceTraders. However applying this approach to this simple environment provides a clear indication of the role which local action-selection plays in hampering attempts to learn SER-optimal behaviour. 

\begin{algorithm}
  \caption{Multiobjective Q($\lambda$) with policy options.}
  \label{algo:moql-options}
  \begin{algorithmic}[1]
    \Statex input: learning rate $\alpha$, discounting term $\gamma$, eligibility trace decay term $\lambda$, number of objectives $n$, action-selection function $f$ and any associated parameters, set of policy options $P$
    \For {all states $s$, policy-options $p$ and objectives $o$}
    	\State initialise $Q_o(s,p)$
    \EndFor
    \For {each episode} 
        \For {all states $s$ and options $p$}
    		\State $e(s,p)$=0
        \EndFor
    	\State observe initial state $s_t$
        \State select option $p_e$ using $f(Q(s_t))$ (with possible exploratory selection)
        \State select $a_t$ from $p_e(s_t)$
        \For {each step of the episode}
 			\State execute $a_t$, observe $s_{t+1}$ and reward $R_t$
 			\State select $a^\prime$ from $p_e(s_{t+1})$
            \State $\delta = R_t + \gamma Q(s_{t+1},p_e) - Q(s_t,P_e)$
            \State $e(s_t,p_e)$ = 1
            \For {each state $s$}
            	\State $Q(s,p_e) = Q(s,p_e) + \alpha\delta e(s,p_e)$
                \State $e(s,p_e) = \gamma \lambda e(s,p_e)$
            \EndFor
            \State $s_t = s_{t+1}, a_t = a^\prime$
    	\EndFor
    \EndFor
  \end{algorithmic}
\end{algorithm}

\subsection{Policy options experimental results}
Table \ref{tab:Space-Traders-Option} shows the distribution of the final policy learned using policy options on the original SpaceTraders environment in comparison to the other methods studied earlier in the paper. It can be seen that the most common result (14/20 runs) is the DI policy, which is the desired optimal policy, but several different sub-optimal policies occur in some trials. While these results represent a substantial improvement on the baseline MOQ-learning method, which only converges to DI in one of twenty trials, they are still well short of the ideal outcome which would always converge to DI.

\begin{table}[hbt!]
    \centering
    \begin{tabular}{@{}|c|c|c|c|c|c|c|@{}}
    \hline
    Policy & DI & ID & II & IT & TI & DD\\ \hline
    Baseline & \textcolor{red}{1} & 13 & 4  & 2  & 0 & 0\\ \hline
    Reward Design & \textcolor{red}{10} & 5 & 1 & 1 & 2 & 1\\ \hline
    MOSS & \textcolor{red}{15} & 0 & 0 & 3 & 2 & 0\\ \hline
    Policy Options - constant learning rate& \textcolor{red}{14} & 2 & 1 & 1 & 2 & 0\\ \hline
    Policy Options - decayed learning rate& \textcolor{red}{20} & 0 & 0 & 0 & 0 & 0\\ \hline
    \end{tabular}%
    \caption[twenty independent runs of the Algorithm \ref{algo:moql-options}]{The final greedy policies learned in twenty independent runs of the Policy Options MOQ-Learning algorithm for the Space Traders environment, with both a constant and decayed learning rate - red indicates the SER-optimal policy. Results from earlier methods are also shown as a basis for comparison.}
    \label{tab:Space-Traders-Option}
\end{table}

\subsection{Policy options and noisy estimates}

Given that the policy-options method eliminates any local decision-making, it should theoretically always converge to the SER-optimal policy. The failure to do so suggests that the \emph{noisy estimates} issue plays a critical role in causing these variations in final policy. The results in the final row of Table \ref{tab:Space-Traders-Option} confirm this, as when run in conjunction with a decayed learning rate, the policy-options method converges to the correct policy in all twenty trials.

The results in earlier sections showed that noisy estimates can impair the ability of value-based MORL algorithms to stably converge to a consistent policy. However these earlier results didn't cleanly isolate the impact of noisy estimates, as those algorithms could also fail due to the underlying limitations of local action-selection in finding SER-optimal policies. In contrast, the policy options algorithm entirely eliminates the use of local action-selection, and so provides an opportunity to more clearly determine the extent to which the noisy estimates issue interferes with the learning of optimal policies in stochastic environments. Therefore in this section we provide a more detailed analysis of the behaviour of policy options MOQ-learning, under both a constant learning rate (where noisy estimates are an issue) and a decaying learning rate (where the noise in the estimates disappears in the later stages of learning).

\begin{figure}[hbt!]
     \centering
     \begin{subfigure}[h]{0.45\textwidth}
         \centering
         \includegraphics[width=\textwidth]{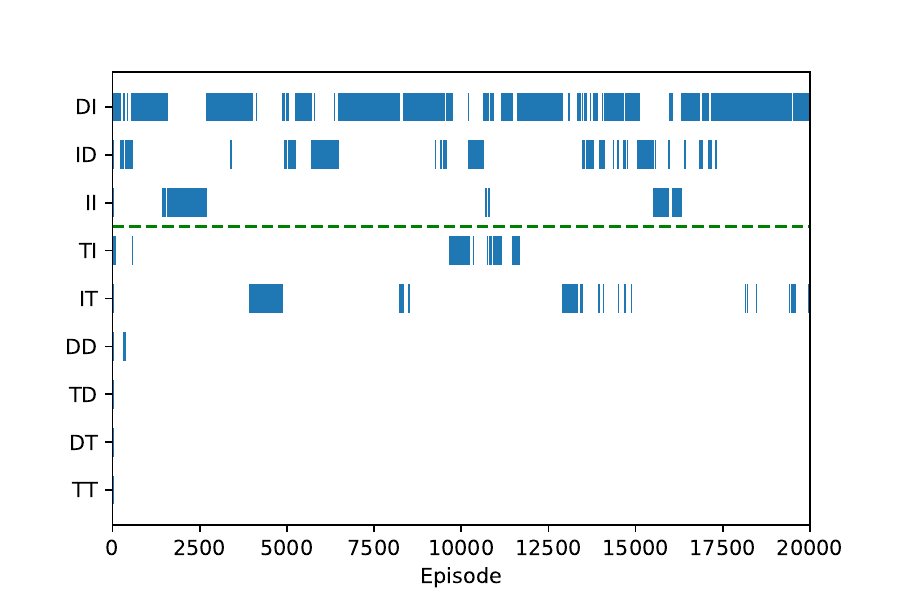}
         \caption{Policy DI}
         \label{fig:Option DI}
     \end{subfigure}
     \begin{subfigure}[h]{0.45\textwidth}
         \centering
         \includegraphics[width=\textwidth]{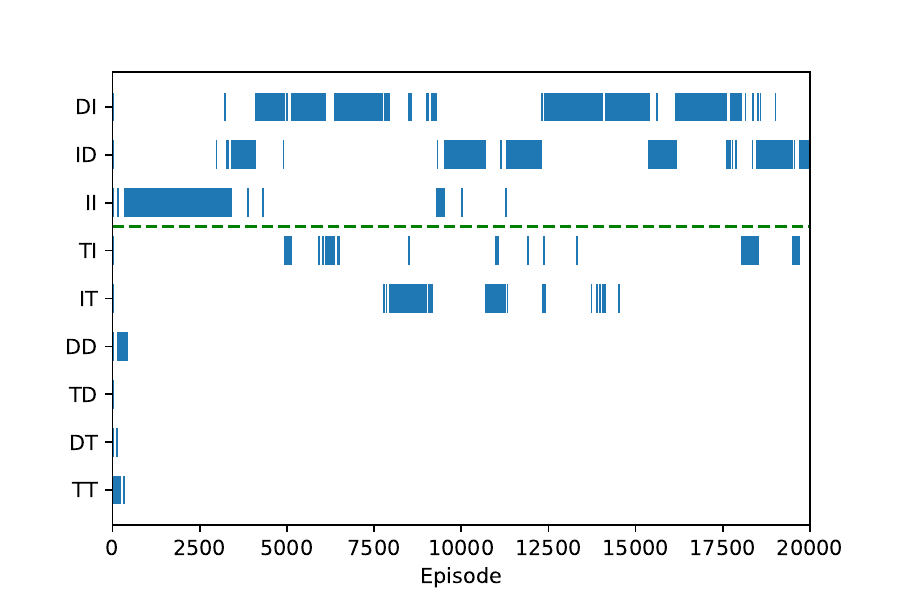}
         \caption{Policy ID}
         \label{fig:Option ID}
     \end{subfigure}
     \begin{subfigure}[h]{0.45\textwidth}
         \centering
         \includegraphics[width=\textwidth]{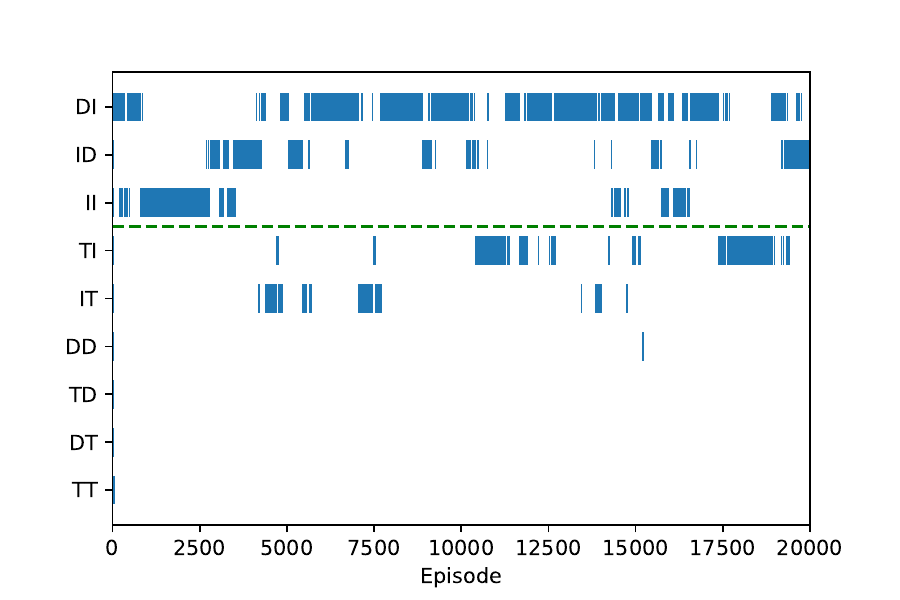}
         \caption{Policy II}
         \label{fig:Option II}
     \end{subfigure}
          \begin{subfigure}[h]{0.45\textwidth}
         \centering
         \includegraphics[width=\textwidth]{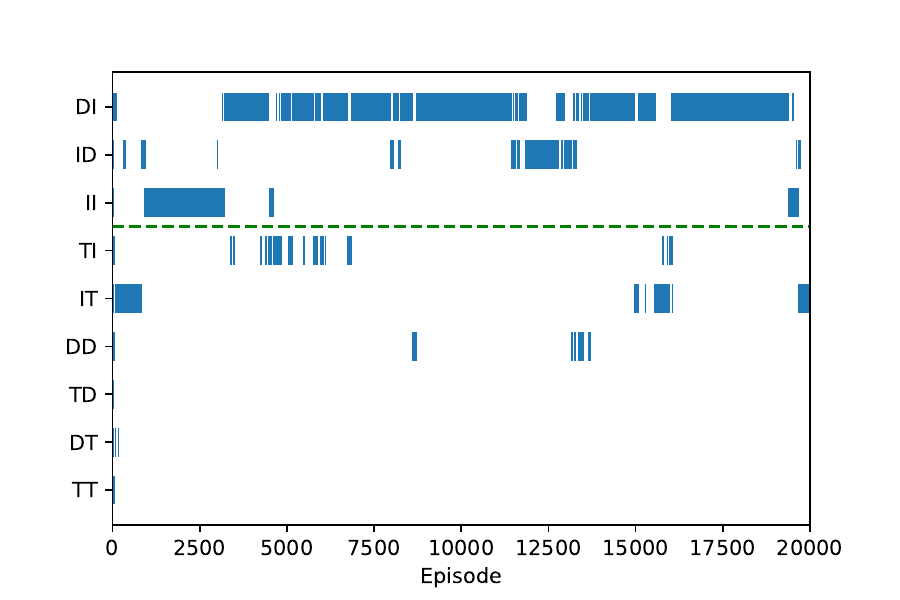}
         \caption{Policy IT}
         \label{fig:Option IT}
     \end{subfigure}
     \begin{subfigure}[h]{0.45\textwidth}
         \centering
         \includegraphics[width=\textwidth]{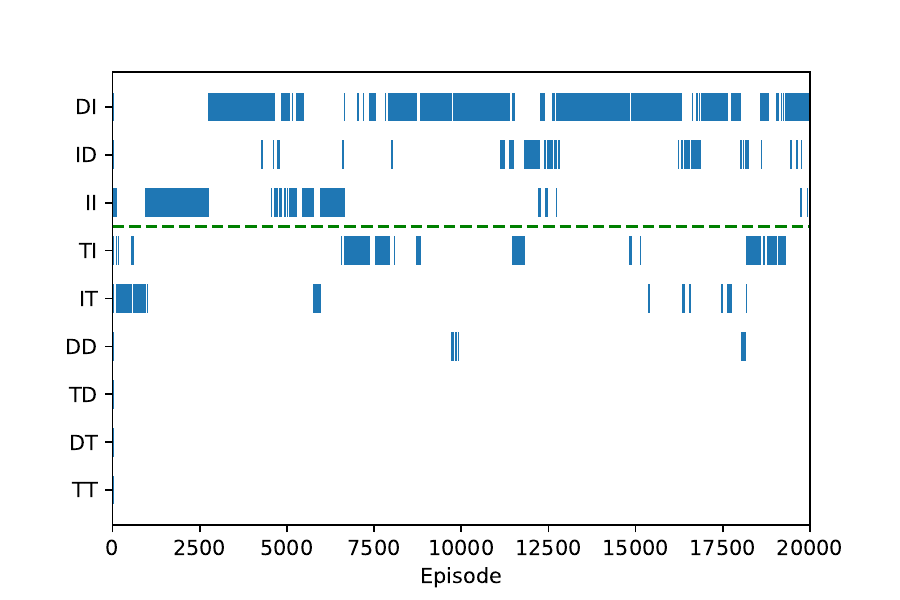}
         \caption{Policy TI}
         \label{fig:Option TI}
     \end{subfigure}
     \caption[Policy charts for Option Learning]{Policy charts for five sample runs of the Policy Options MOQ-Learning algorithm on Space Traders}
     \label{fig:Option Policy Charts}
\end{figure}

Figure \ref{fig:Option Policy Charts} highlights the noisy estimates issue as it can be seen that the policy identified as being optimal fluctuates on a frequent basis during learning when the learning rate is constant, and quite often includes policies which fail to achieve the threshold on the first objective. Seeing as both the agent's value estimates and action-selection are being performed at the level of complete policies, this can only arise due to errors in those estimated values arising from the stochastic nature of the environment.

\begin{figure}[hbt!]
    \centering
    \includegraphics[width=13.5cm]{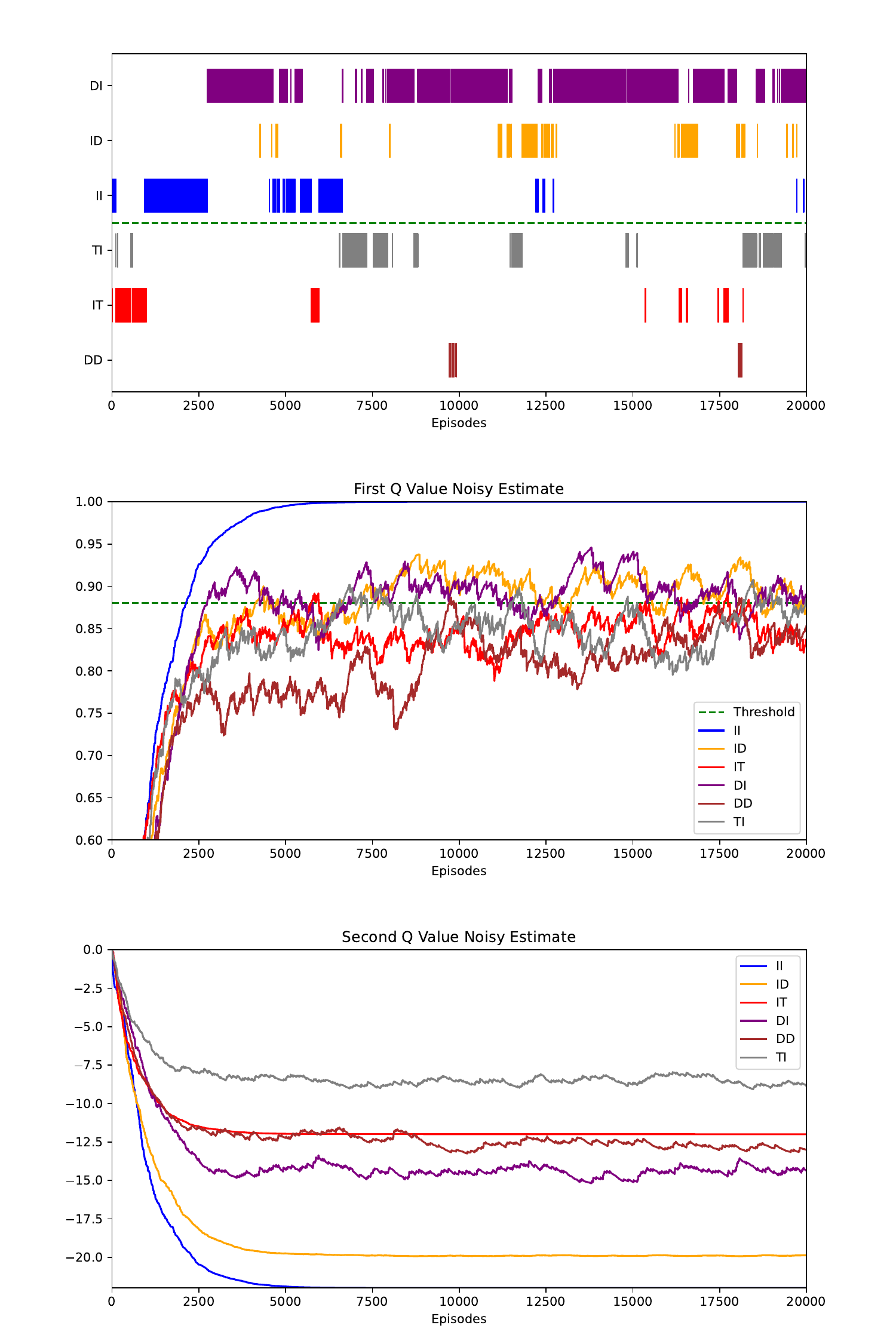}
    \caption[The Noisy Q Value Estimate Issue]{The Noisy Q Value Estimate issue in Policy Options MOQ-learning with a constant learning rate. These graphs illustrate agent behaviour for a single run. The top graph shows which option/policy is viewed as optimal after each episode, while the lower graphs show the estimated Q-value for each objective for each option.}
    \label{fig:NOVE-c}
\end{figure}

\begin{figure}[hbt!]
    \centering
    \includegraphics[width=12cm]{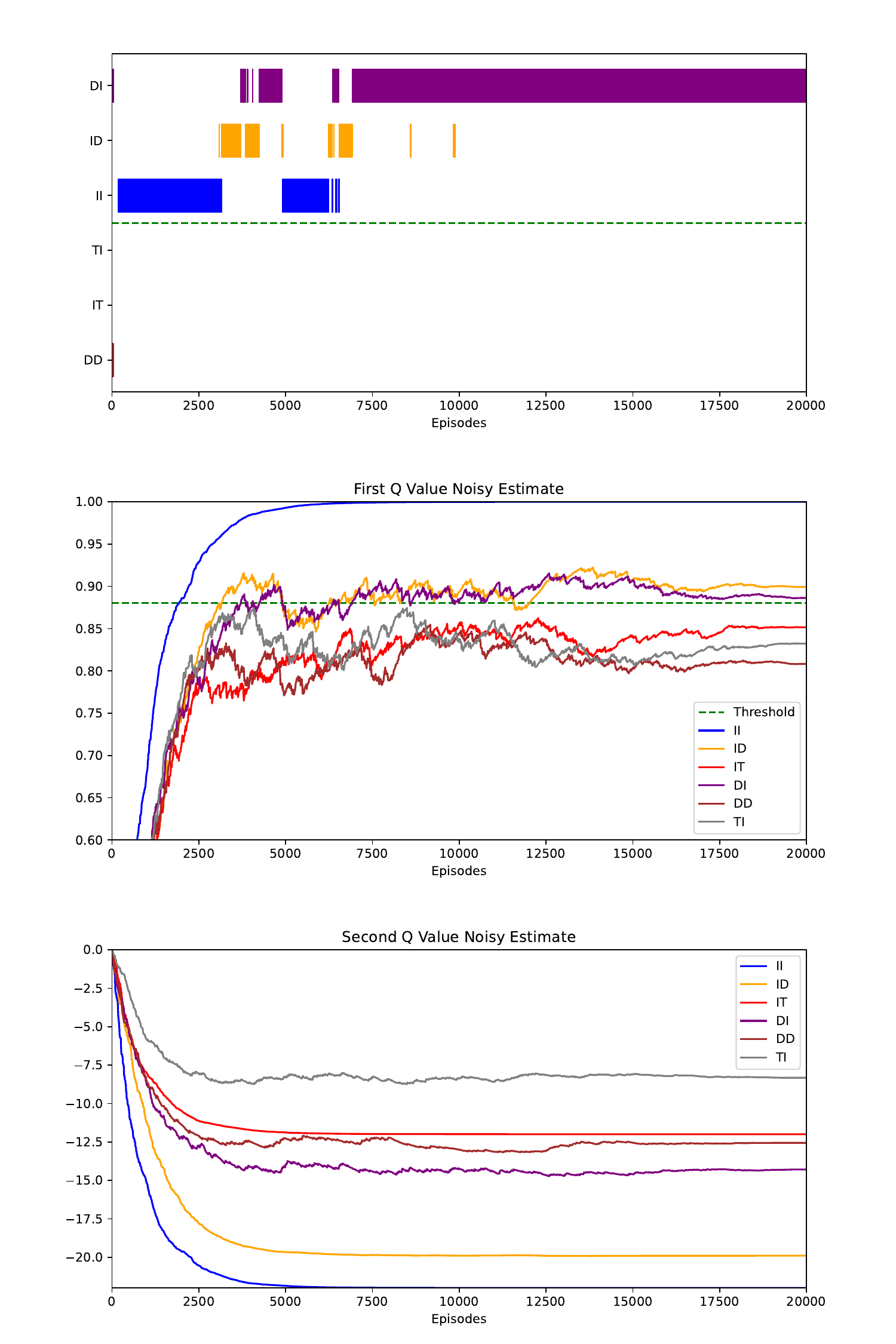}
    \caption[Decay Learning rate in Option learning]{The effect of a decayed learning rate on Policy Options MOQ-learning. Notice the increased stability of both the Q-values and option selection in the later episodes when the learning rate has decayed to a small value.}
    \label{fig:NOVE-d}
\end{figure}

Figure \ref{fig:NOVE-c} visualises a single trial of policy options for the original Space Traders problem which eventually selects policy TI. The first layer is the normal policy chart where each policy has been assigned a unique colour for clarity. The middle layer indicates the Q-value at state A for the first objective, and the bottom layer shows the Q-value at state A for the second objective. As we can see from these three graphs, due to the combination of the stochastic environment and hard-coded threshold, the optimal policy never stabilised even though it stays on DI the desired optimal policy most of time. There is an extreme case just before 10,000 episodes and again at around 18,000 episodes, where the policy DD (in brown color) has several unlikely successes in a row, and its estimated value rises above the threshold. This means it is temporarily identified as the optimal policy in the policy chart at those times, despite in actuality being the least desirable policy -- this indicates the extent of the impact of noisy estimates within a highly stochastic environment such as Space Traders.

In contrast Figure \ref{fig:NOVE-d} is a representative trial of the policy options algorithm using a decayed learning rate. As we can see from the top policy chart, this time the agent did converge to the desired optimal policy DI and stabilised on this solution after around 10,000 episodes. The decaying learning rate reduces the influences of the occasional unsuccessful or successful runs during later learning episodes, meaning the agent is much less likely to switch its choice of optimal policy (and in particular, far less likely to incorrectly identify a policy which fails to meet the threshold for the first objective as being optimal).

When combined with decaying the learning rate, policy options learning is able to address both the local decision-making issue and the problem of noisy Q-value estimates for stochastic environments. However this method suffers from a more fundamental problem – the curse of dimensionality. For problems with more states and actions, the number of pre-defined options are going to increase exponentially, and so this method is not able to scale up to solve more complex problems in real-life. 

\section{Conclusion}

Multi-objective Q-learning is an extension of scalar value Q-learning that has been widely used in the multi-objective reinforcement learning literature. However it has been shown to have limitations in terms of finding the SER-optimal policy for environments with stochastic state transitions. This research has provided the first detailed investigation into the factors that influence the frequency with which value-based MORL Q-learning algorithms learn the SER-optimal policy under a non-linear scalarisation function for a stochastic state environment.

\subsection{Major Findings}
This study explored three different approaches to address the issues identified by \cite{VamplewEnvironmental2022} regarding the inability of multiobjective Q-learning methods to reliably learn the SER-optimal policy for environments with stochastic state transitions. The first approach was to apply reward design methods to improve the MORL agent's performance in stochastic environments. The second approach (MOSS) utilised global statistics to inform the agent's action selection at each state. The final approach was to use policy options (options defined at the level of complete policies). 

The results for the first approach showed that with a new reward signal which provided additional information about the probability of transitioning to terminal states, standard MOQ-learning was able to find the desired optimal policy in the original Space Traders problem. However, using a slightly modified variant of Space Traders we demonstrated that in general it may be too hard or even impossible to design a suitable reward structure for any given MOMDP.

It was found in the second approach that the augmented state combined with use of global statistics in the MOSS algorithm clearly outperformed the baseline method for the Space Traders problem. However, the MOSS algorithm fails to find the correct policy for the Space Traders ID variation of the environment, and therefore does not provide a reliable solution to the task of finding SER-optimal policies.

The results for the third approach reveal that policy option learning is able to identify the optimal policy for the SER criteria for a relatively small stochastic environment like Space Traders. However clearly this method still fails from a more fundamental problem – the curse of dimensionality means it is infeasible for environments containing more than a small number of states and actions. 

The key contribution of this work is isolating the impact of noisy Q-estimates on the performance of MO Q-learning methods. The final experiments using policy options clearly illustrated the extent to which noisy estimates can disrupt the performance of MO Q-learning agents. By learning Q-values and performing selection at the level of policies rather than at each individual state, this approach avoids the issues with local decision-making which are the primary cause of difficulty in learning SER-optimal policies \cite{bryceProbabilisticPlanningMultiobjective2007,VamplewEnvironmental2022}. However the empirical results showed that when used in conjunction with a constant learning rate, the variations in estimates introduced by the stochasticity in the environments lead to instability in the agent's greedy policy. In many cases this meant the final greedy policy found at the end of learning was not the SER-optimal policy.  Further experiments revealed that using a decayed learning rate was able to mitigate the effect of environmental stochasticity and converge to the correct final policy. 

These results highlight that in order to address the problems of value-based MORL methods on stochastic environments, it will be essential to solve both the local decision-making issue and also the noisy estimates issue, and that a decaying learning rate may be valuable in addressing the latter.

\subsection{Future work}
There are two issues existing for MOQ-learning in stochastic environments (the core stochastic SER issue caused by local decision-making and also the noisy Q value estimates), therefore a successful algorithm must address both of those problems together. Due to the flaws in each investigated method, none of them could be directly applied into real-world applications, and so there is a need for further research to develop more reliable approaches for SER-optimal MORL.

The first recommendation for future research is to look at policy-based methods such as policy gradient RL. As these methods directly maximise the policy as a whole by defining a set of policy parameters, therefore they do not have the local decision-making issue faced by model-free value-based methods such as MOQ-learning. Several researchers have developed and assessed policy-based methods for multi-objective problems \cite{parisi2014policy} \cite{bai2021joint}. However most policy-based MORL methods produce stochastic policies, whereas in some applications deterministic policies may be required. So these algorithms may require modification in order to deal with this constraint.

A second promising research direction is to investigate \acrfull{DRL}. Conventional value-based RL learns a single value per state-action pair which represents the expected return. Distributional reinforcement learning on the other hand works directly with the full distribution of the returns instead. This can be beneficial for MORL, as shown by \cite{hayesmulti2022} who applied  Distributional Multi-objective Value Iteration to find optimal policies for the ESR criteria. The additional information about the rewards captured by DRL algorithms could potentially prove useful in overcoming both the noisy estimates and stochastic SER issues.


%
\section*{Conflict of interest}
The authors declare that they have no conflict of interest.

\bibliographystyle{spbasic}      
\bibliography{stochastic.bib}   

\end{document}